\documentclass[letterpaper, 10 pt, conference]{ieeeconf}  

\IEEEoverridecommandlockouts                              
\usepackage[table]{xcolor}
\usepackage{graphicx}				
\usepackage[labelformat=simple]{subcaption}	
\usepackage{amssymb}
\usepackage{hyphenat,fancyref,hyperref,xspace,tabularx}
\usepackage{adjustbox}
\usepackage{makecell}
\usepackage{wrapfig}
\usepackage{amsmath}
\usepackage{mathtools}
\usepackage{listings}
\usepackage{url}
\usepackage{epstopdf}
\usepackage{array}
\usepackage{tikz}
\usepackage{amsmath}
\usepackage{algorithm}
\usepackage{algpseudocode}
\usepackage{multirow}
\usepackage{flexisym}
\usepackage{amsmath}
\usepackage{array}
\usepackage{bbold}
\usepackage{subfiles}
\usepackage{cite}
\usepackage{gensymb}
\usepackage{soul}
\usepackage[normalem]{ulem}
\usepackage[margin=0pt,font=small,labelsep=period]{caption}
\usepackage{calc,booktabs,xcolor,url,hyperref}

\definecolor{green}{RGB}{0,255,0}
\definecolor{dark_green}{RGB}{25,155,25}

\newcommand{\systemname}{{\it AuraSense}\xspace}

\title{\systemname: Robot Collision Avoidance by Full Surface Proximity Detection}

\author{Xiaoran Fan, Riley Simmons-Edler, Daewon Lee, \\ Larry Jackel, Richard Howard and Daniel Lee
\thanks{All authors are with Samsung AI Center NY, 837 Washington Street New York, New York 10014
}%
}

\begin{document}

\maketitle
\thispagestyle{empty}
\pagestyle{empty}

\begin{abstract}
Perceiving obstacles and avoiding collisions is fundamental to the safe operation of a robot system, particularly when the robot must operate in highly dynamic human environments. Proximity detection using on-robot sensors can be used to avoid or mitigate impending collisions. However, existing proximity sensing methods are orientation and placement dependent, resulting in blind spots even with large numbers of sensors. In this paper, we introduce the phenomenon of the Leaky Surface Wave (LSW), a novel sensing modality, and present \systemname, a proximity detection system using the LSW. \systemname is the first system to realize no-dead-spot proximity sensing for robot arms. It requires only a single pair of piezoelectric transducers, and can easily be applied to off-the-shelf robots with minimal modifications.  We further introduce a set of signal processing techniques and a lightweight neural network to address the unique challenges in using the LSW for proximity sensing. Finally, we demonstrate a prototype system consisting of a single piezoelectric element pair on a robot manipulator, which validates our design. We conducted several micro benchmark experiments and performed more than 2000 on-robot proximity detection trials with various potential robot arm materials, colliding objects, approach patterns, and robot movement patterns. \systemname achieves 100\% and 95.3\% true positive proximity detection rates when the arm approaches static and mobile obstacles respectively, with a true negative rate over 99\%, showing the real-world viability of this system.
\end{abstract}

\section{Introduction}
\label{s:intro}
\vspace{-0.05in}

As robots and robot manipulators work in dynamic environments, unexpected collisions with people and obstacles must be avoided. A robot colliding with the environment can damage itself or its surroundings, and can harm humans in the workspace. Collision avoidance systems enable the robot to detect approaching obstacles in proximity to the robot before collision and take measures to avoid or mitigate impact. As such, there has been extensive research on collision avoidance systems for robotic manipulators.

Unlike collision avoidance systems for automobiles, robot manipulators usually operate in confined spaces, where collision avoidance depends on accurate short range sensing in cluttered 3D environments.
Many existing collision avoidance methods use cameras and computer vision-based object recognition or 3D shape reconstruction to detect and react to obstacles. However, these approaches have several limitations-- their performance suffers when faced with obstacle occlusions, poor light conditions, and transparent or mirrored objects that are hard to detect visually. Further, camera-based approaches are typically not accurate over very short ranges ($\leq 10$cm), depending on camera focal length, and any single camera has a limited field of view.

\begin{figure}[t]
\centering
{\includegraphics[width=1\columnwidth]{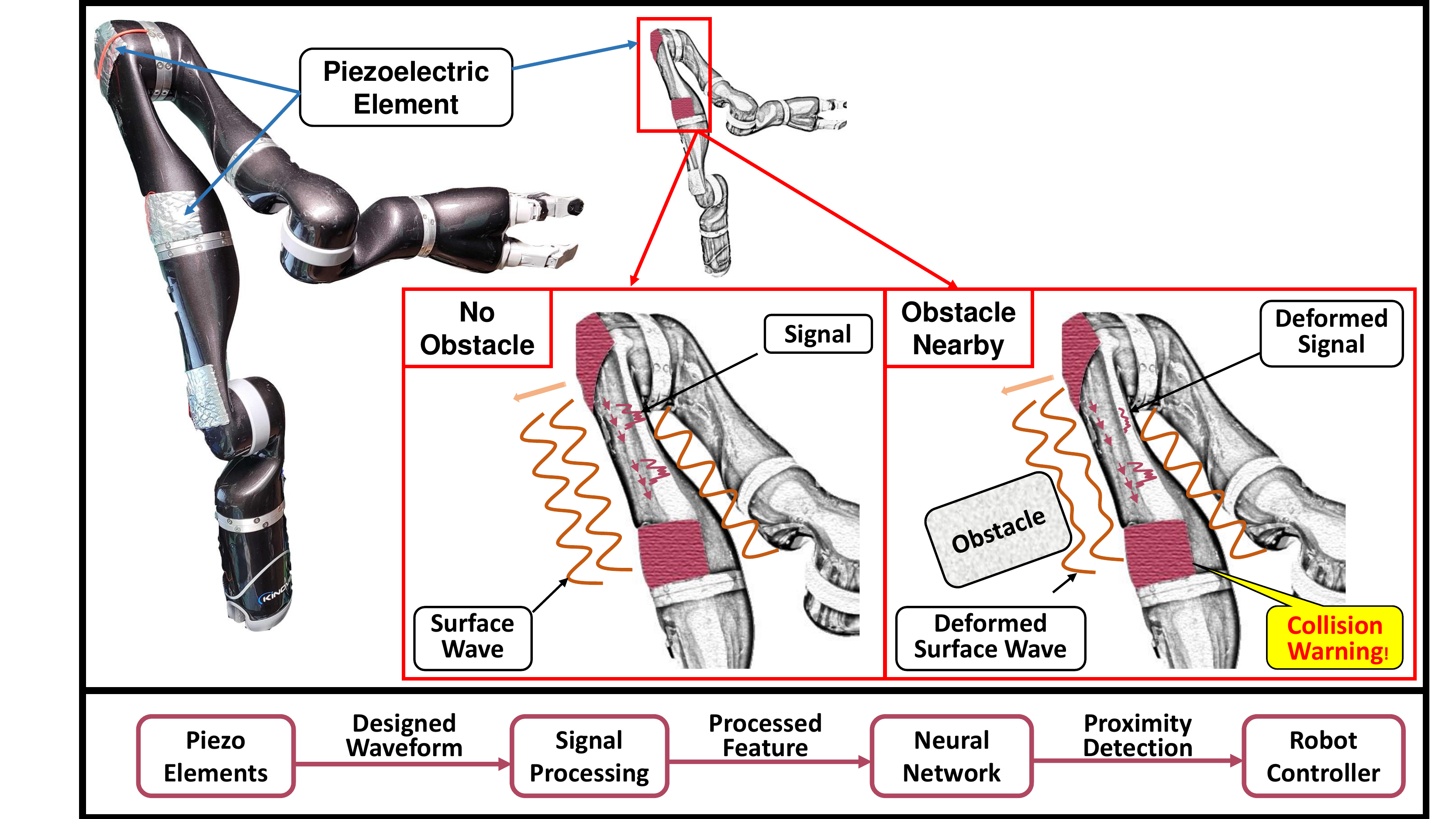}}
\caption{\systemname system configuration and overview.}
\label{fig:fig1}
\vspace{-0.15in}
\end{figure}

To address this need for short range detection, proximity sensors such as ultrasonic proximity sensors~\cite{cho2017compliant}, millimeter wave radar~\cite{geiger2019160}, infrared proximity sensors~\cite{benet2002using}, and short range LiDAR~\cite{amzajerdian2016imaging} have been proposed specifically for robot collision avoidance. These methods also have limitations: LiDAR and millimeter wave radar are expensive, and all these methods are all highly directional-- Effective coverage requires multiple sensors distributed throughout the robot, and blind spots can be difficult to eliminate entirely without vast numbers of sensors. This complicates robotic system design and adds significant sensor management overhead.

In this paper, we propose a novel sensing modality, which we term the {\it Leaky Surface Wave} (LSW), that enables no-dead-spot short range proximity detection for robots. We describe a proximity detection system using this principle, \systemname, which is lightweight, economical, can be attached to an off-the-shelf robotic manipulator or mobile robot body with minimal modifications, and provides proximity detection of all objects with sufficient cross-sectional area across the entire surface of the robot. \systemname is to our knowledge the first system that can perform full surface and omnidirectional on-robot proximity detection for an entire linkage using only a single sensor pair.


\systemname uses one or more piezoelectric elements attached to a robot arm, and transmits excitation signals from one element through the robot arm to other elements.
This acoustic energy travels through the whole surface of the robot arm, which, through its vibration, couples the signal into the air. This acoustic signal decays rapidly with distance from the robot surface, forming an “aura" that is disturbed by the presence of a foreign object in proximity. An approaching obstacle which enters this aura will establish a standing wave pattern between the obstacle and the surface of the robot, changing the acoustic impedance of the system. This change can be measured by the source transducer or another attached elsewhere on the arm, potentially at a point far from the obstacle, allowing us to perform proximity detection.

To realize this concept, we address a number of technical and implementation challenges. First, the major component of the signal is received from the surface of the robot rather than the leaky over-the-air signal. However, only the leaky over-the-air signal contains information useful for proximity detection. We employ a set of hardware enhancements and signal processing techniques to extract this minor leaky signal from the large surface signal. 
Second, the robot arm itself introduces both mechanical and electrical noise which is received by the attached piezoelectric element. We solve this issue by customizing the waveform, and further digitally filtering the noise. Last but not least, the received signal will vary non-linearly depending on the robot pose (including self-detection) and relative obstacle position/velocity as the robot moves around. 
To resolve these issues, we use a lightweight one-dimensional convolutional neural network to identify whether a given received audio sequence corresponds to the presence of a non-self obstacle.

We present an end-to-end proximity detection system with a pair of low cost ($<$ \$2 ) piezoelectric elements attached to an off-the-shelf commercial robot arm, and demonstrate no-dead-spot proximity sensing with several practical experiments. The simplicity of this design makes it easy to embed into a robot with minimum modifications.
To summarize, this paper makes the following contributions:
\begin{itemize}
    \item We introduce a novel sensing modality, the Leaky Surface Wave, which turns the entire robot surface into a tactile skin that allows no-dead-spot proximity sensing on a robot arm. We build a real-world system, \systemname, to demonstrate this new sensing capability.
    \systemname is the first system to realize whole surface collision avoidance on a robot linkage. 
    \item We explore the physics behind this new sensing modality in the context of robotic systems. We propose several signal processing algorithms, a hardware configuration, and a lightweight 1D-CNN algorithm to specifically address the unique challenges in our scenario. 
    \item We implement an \systemname prototype on a 7 degree-of-freedom (DOF) manipulator with a 96 kHz sampling rate audio chain, and conduct comprehensive evaluations of its real-world performance. Our multi-scenario experiments test performance on various approaching objects with different dielectric and mechanical properties, a range of approaching speeds and angles, and random robot movement patterns. \systemname attains a true positive rate of over 95\% with a less than 1\% false positive rate for on-robot proximity detection.

\end{itemize}
\vspace{-0.05in}

\section{Related work}
\label{s:rel}
\vspace{-0.05in}
The LSW bares some similarity to surface acoustic waves (SAWs)~\cite{hashimoto2000surface}, which also propagate waves along an object's surface between transmitter and receiver. However, SAWs rely on the object itself being piezoelectric, operate at much higher frequencies (megahertz range), don't leak waves into the air, and are mostly used as signal filters.

On the application side, existing systems for on-robot collision avoidance can be categorized into three types: 1) Computer vision-based collision detection, 2) Time-of-Flight (ToF) sensor based, and 3) Interference and capacitive sensing based. 
Computer vision-based solutions using monocular cameras~\cite{alvarez2016collision}, stereo cameras~\cite{chen2020bio}, and depth cameras~\cite{schmidt2014depth} have been proposed, which seek to recognize obstacles in the camera view. However, even with high performance neural network models and algorithms, computer vision-based solutions are still limited by camera viewing angle, and lighting conditions, and may perform poorly on reflective and transparent obstacles. Camera based solutions are also in general not well-suited for short range ($\leq$ 10cm) sensing, as such short focal lengths can require specialized lenses to avoid heavy distortion. 
Next, ToF methods can also be used to help robots avoid nearby obstacles. ToF signals can be acquired from phased radar~\cite{fan2020towards}, lidar~\cite{amzajerdian2016imaging}, or ultrasound sensors~\cite{cho2017compliant}. However, 
tens or even hundreds of ToF sensors may be needed to provide whole surface proximity detection without blind spots. This is impractical due to 1) the high cost of ToF sensors such as lidar or millimeter wave radar, and 2) large numbers of of ToF sensors emitting at the same time will interfere with each other, which complicates system design. 
Finally, other type of sensing modalities such as infrared sensors~\cite{benet2002using} or capacitive coupling sensors~\cite{poeppel2020robust} which leverage signal interference patterns have been proposed for on-robot collision avoidance. However, these sensors are orientation-dependent, and capacitive sensing only works on obstacles that have purposely calibrated dielectric constants, such as water or human tissues. 

\systemname differs practically from these sensing modalities as it enables whole surface sensing capability with only a single transmitter/receiver pair. To our knowledge, \systemname is the first system to provide no-dead-zone proximity sensing among published on-robot proximity sensing systems.  
\vspace{-0.05in}
\section{Methods and System Design}
\label{s:design}
\vspace{-0.05in}

In this section, we first introduce the core of our novel sensing modality. Next, we discuss how to leverage this new sensing modality for on-robot whole surface proximity detection, from robot instrumentation to neural network classification. We include proof-of-concept results throughout to illustrate important aspects of the system design.

\subsection{Leaky Surface Wave}
\vspace{-0.05in}
\label{subsec:phy}
At the core of our sensing method is an acoustic signal, the LSW. By coupling piezoelectric elements to the surface of an object with low acoustic loss, such as a robot arm made of plastic or metal, and applying an excitation signal to the piezoelectric element, acoustic waves propagate within and along the surface of the object. As a result, the surface of the object will vibrate and couple acoustic energy to the air, making the entire surface an acoustic transducer. Notably, piezoelectric elements couple with the surface instead of the air, and could even be embedded within the object.

\begin{figure}
\centering
\begin{minipage}{.485\linewidth}
  \centering
{\includegraphics[width=0.95\textwidth]{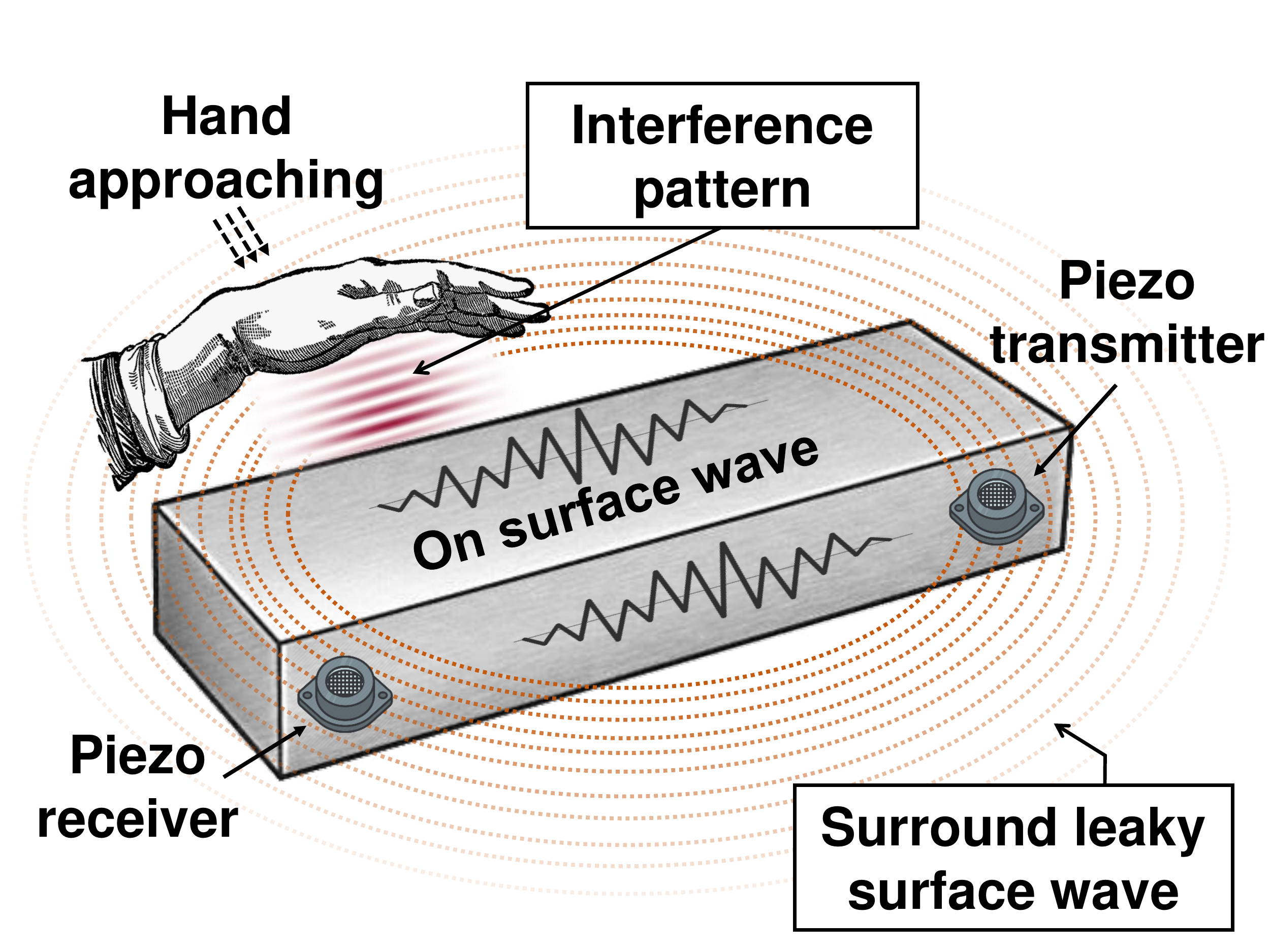}}
\caption{A schematic of the LSW effect. Approaching objects form an interference pattern with the LSW, which affects the surface-guided signal detected by the receiver far from the interference site.}
\label{fig:lsw_schematic}
\vspace{-0.15in}
\end{minipage}%
\hspace{0.1cm}
\begin{minipage}{.485\linewidth}
  \centering
\includegraphics[width=0.95\textwidth]{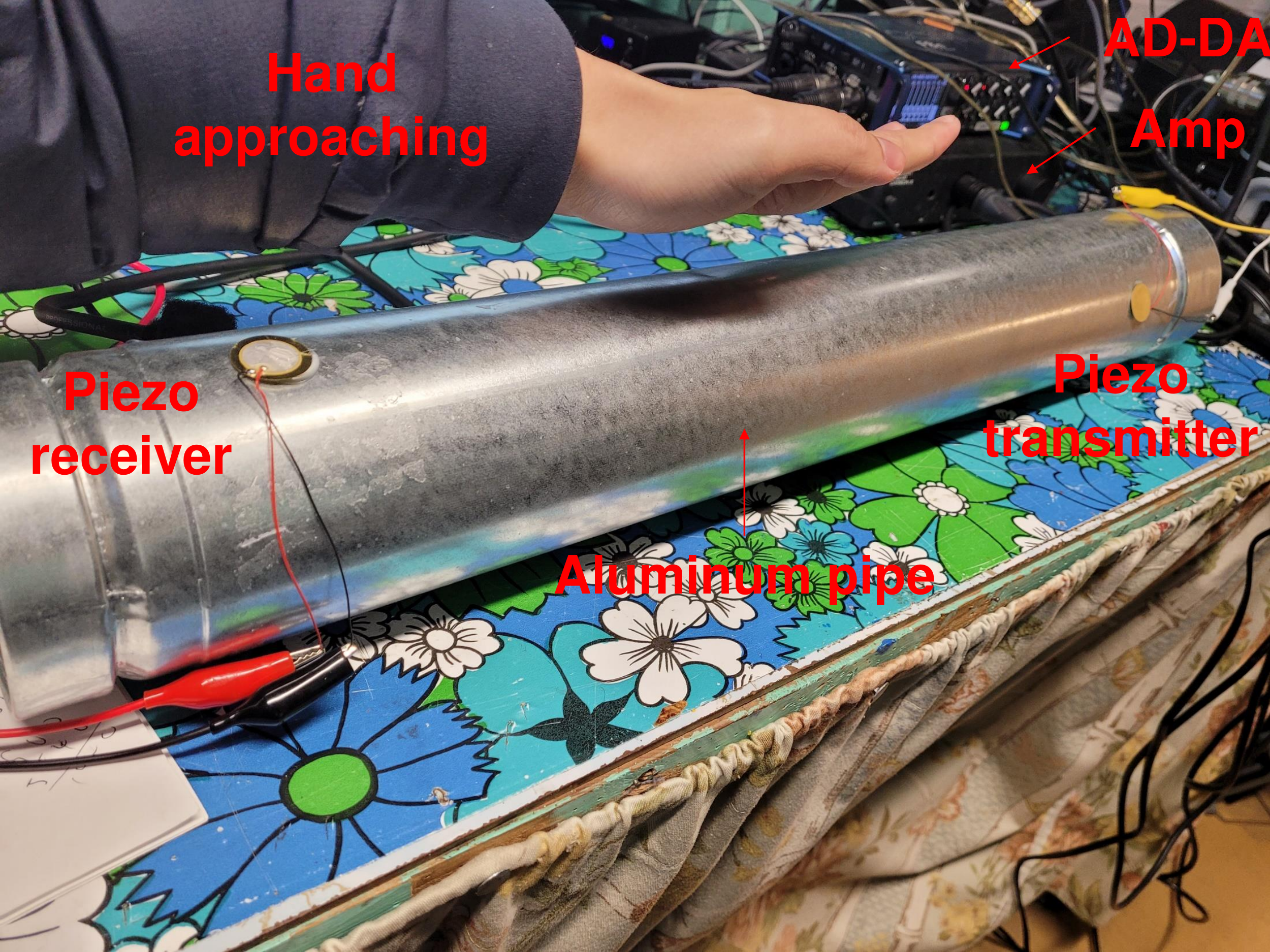}
\caption{ We instrument an aluminum pipe with two piezoelectric elements and apply a continuous sinusoidal excitation signal to one the piezo elements. The other element records the LSW signal.}
\label{fig:sensing_illustration}
\vspace{-0.15in}
\end{minipage}
\end{figure}

A schematic illustrating how the LSW can be distorted is shown in Figure~\ref{fig:lsw_schematic}. While most of the acoustic energy from the transmitter stays on the surface, a small amount ``leaks'' into the air. This leaky signal decays rapidly with distance from the surface, resulting in an acoustic pressure field ``aura'' that surrounds the object. Obstacles close to the surface of the object will establish a standing wave pattern between the obstacle and the object surface
, which perturbs the acoustic pressure field and results in an acoustic impedance change across the entire surface. These changes can be detected by a piezoelectric element elsewhere on or within the object, acting as a receiver. As the signal propagates through the object, obstacles close to any point on the object's surface will cause distortions that can be measured at other points on or within the object, allowing for a single transmitter/receiver pair to detect obstacles close to any part of the coupled object.

\begin{figure}[tb]
\centering
\begin{subfigure}[t]{0.45\linewidth}
\centering
{\includegraphics[width=1.1\textwidth]{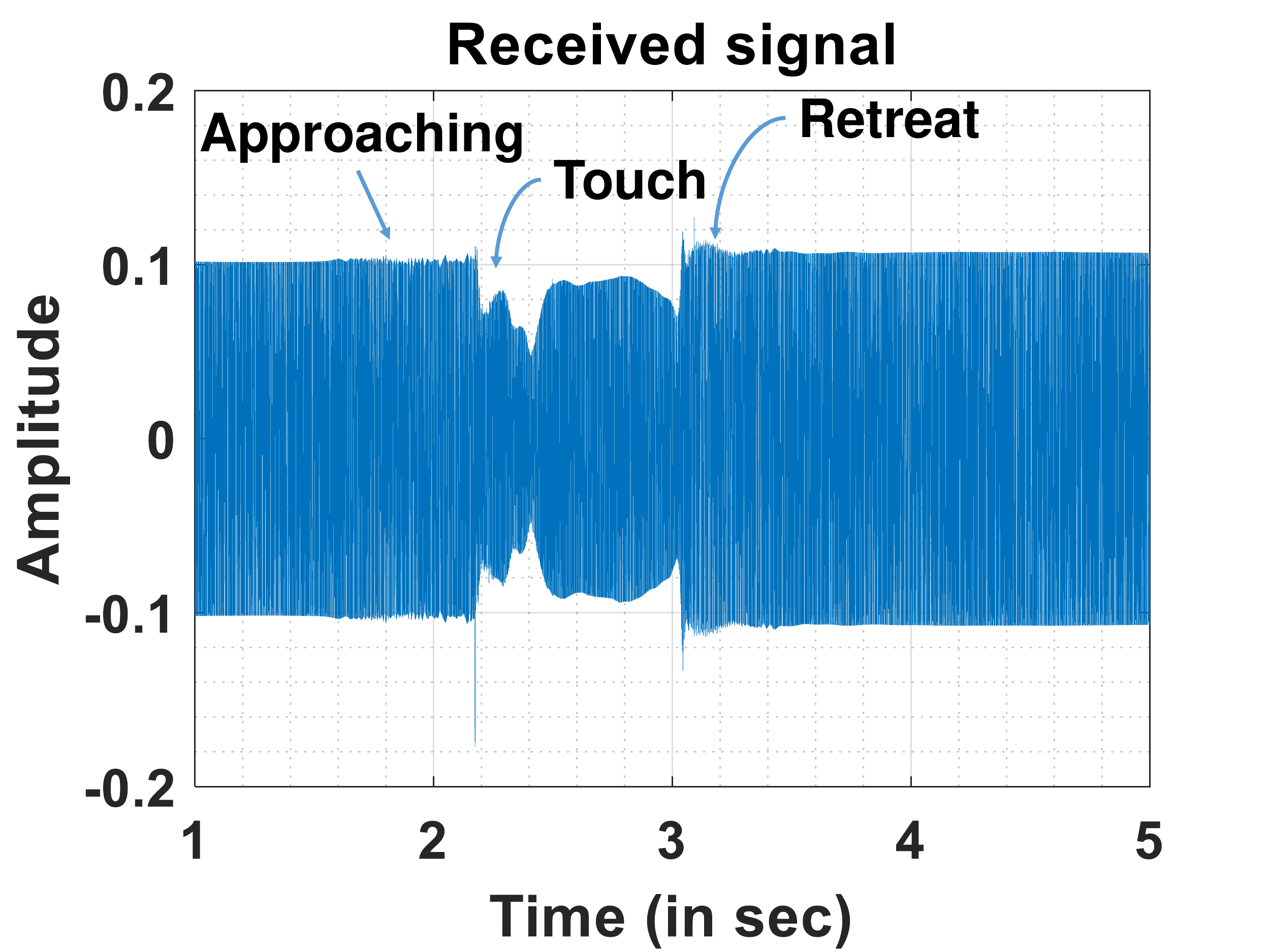}}
\caption{} 
\label{fig:sensing_demo}
\end{subfigure}
\begin{subfigure}[t]{0.45\linewidth}
\centering
{\includegraphics[width=1.1\textwidth]{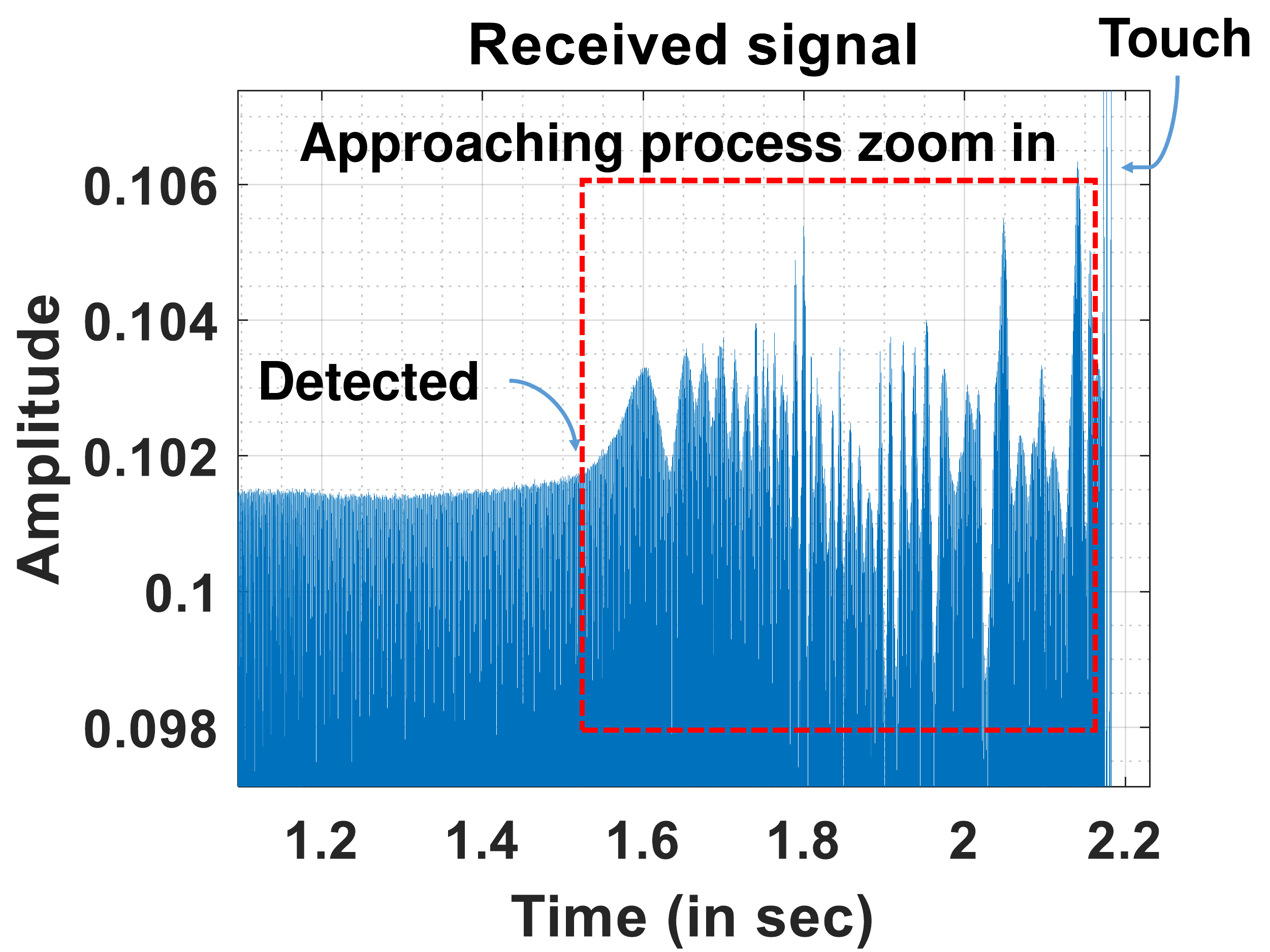}}
\caption{}
\label{fig:sensing_demo_zoom}
\end{subfigure}
\caption{The recorded signal from the first proof-of-concept experiment, showing responses to proximity and contact. (a) The whole signal, including approach, touch and retreat. (b) Enlarged window of when the hand approached the pipe.}
\vspace{-0.2in}
\label{fig:signal_demo}
\end{figure}

\vspace{4pt}\noindent\textbf{Properties of the LSW:}
This surface acoustic pressure field distortion displays a number of useful properties, which we will demonstrate in this section. We performed several simple experiments, using the setup shown in Figure~\ref{fig:sensing_illustration}. We excite the piezo transmitter with a 20 kHz signal. In the first experiment, a person's hand approaches the pipe, touches it, and then retreats. Figure~\ref{fig:sensing_demo} shows the signal recorded in this experiment; touching the pipe (at 2.2-3 seconds) introduces a major signal perturbation. In addition, if we look closely at the signal when the hand is approaching the pipe, shown in more detail in Figure~\ref{fig:sensing_demo_zoom}, the approach of the hand (at 1.5-2.2 seconds) is detected as well. The peaks and dips reflect the standing wave pattern between the hand and the surface, which have a peak-to-peak distance of $d=\lambda/2$, where $\lambda$ is the wavelength of the signal in air. Importantly, the amplitude of the distortion increases as the hand gets closer to the pipe. This signal pattern varies depending on the nature of the approaching obstacle and the obstacle's position and velocity relative to the surface. While these distortions can be small and hard to interpret, measuring them allows the LSW to be used for diverse sensing applications, including proximity detection, object recognition, and tactile sensing.

\begin{figure}[tb]
\centering
\begin{subfigure}[t]{0.45\linewidth}
\centering
{\includegraphics[width=1.1\textwidth]{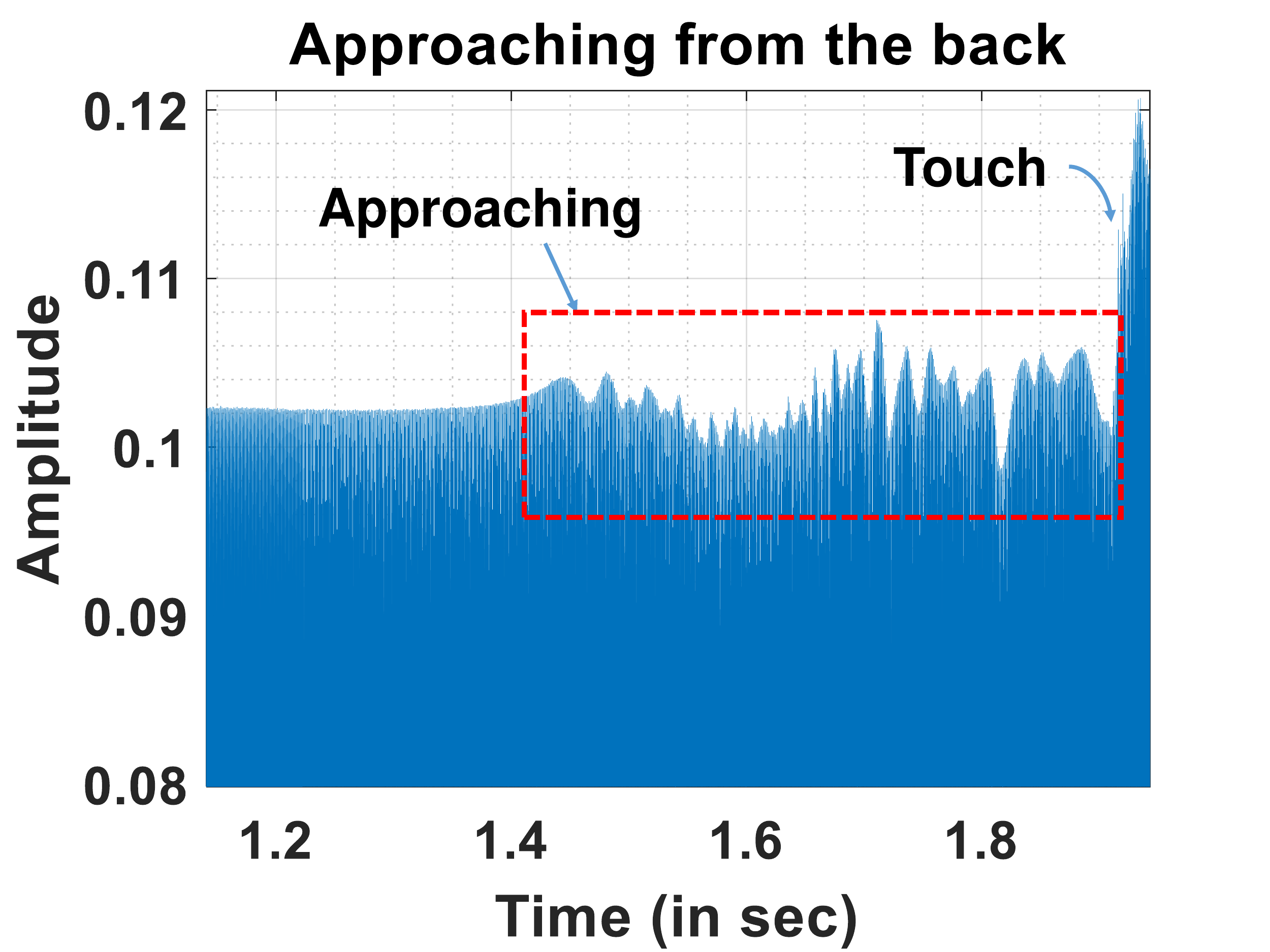}}
\caption{} 
\label{fig:back}
\end{subfigure}
\begin{subfigure}[t]{0.45\linewidth}
\centering
{\includegraphics[width=1.1\textwidth]{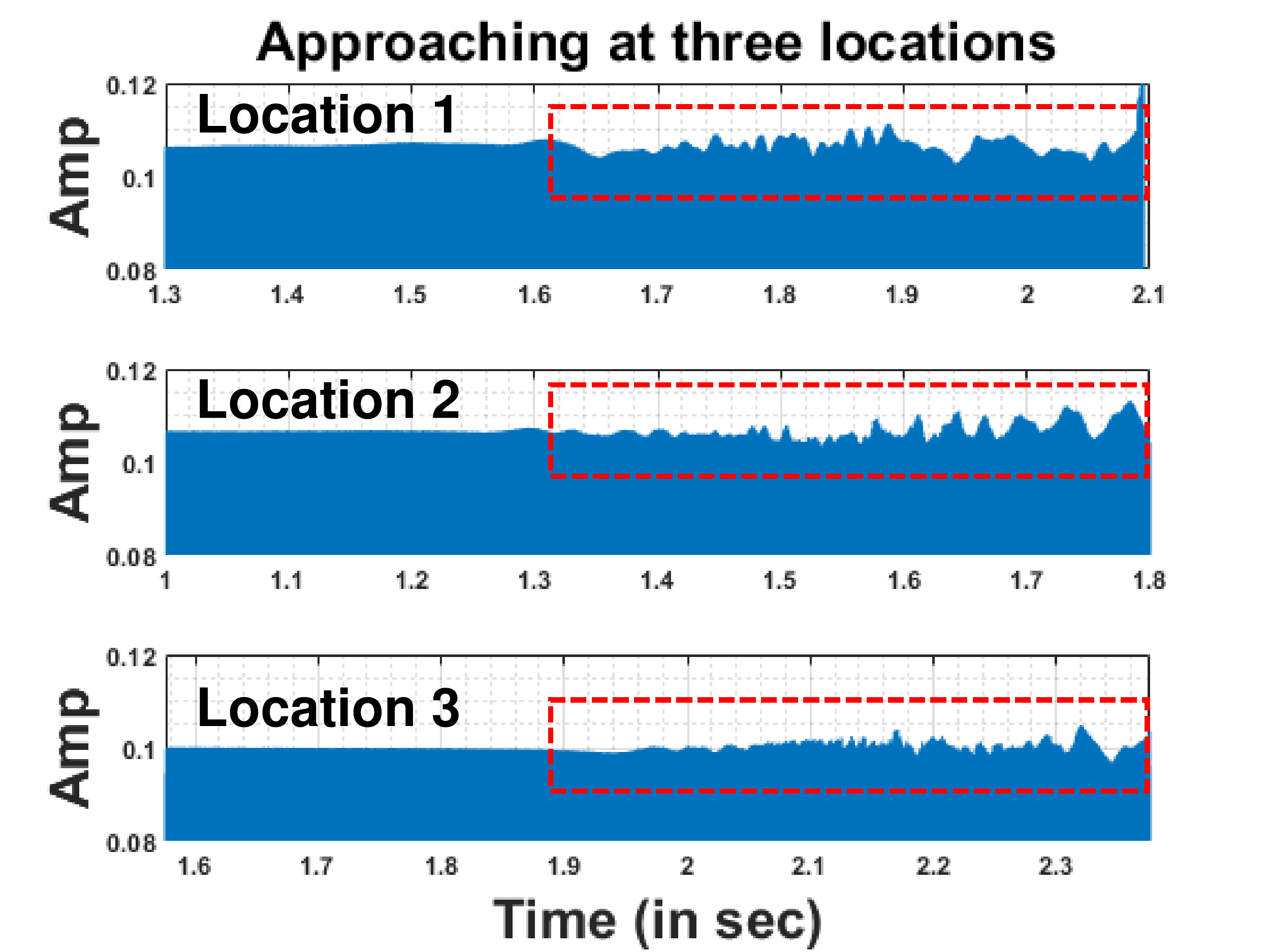}}
\caption{}
\label{fig:diff_loc}
\end{subfigure}
\caption{The signal recorded when the test pipe is approached from the side opposite the receiver and transmitter. The hand's approach is detected without line of sight to either piezo element. (a) The signal as a hand approaches the back side of the aluminum pipe. (b) Hand approach signals from approaching three different locations on the pipe.}
\vspace{-0.2in}
\label{fig:insights}
\end{figure}

Next we demonstrate how the LSW signal allows for detection across whole surface. We performed an experiment where we mounted piezo elements on one side of an aluminum pipe, while a human hand approaches from the opposite side. The resulting signal is shown in Figure~\ref{fig:back}. The proximity signal pattern is clearly visible. Further, in Figure~\ref{fig:diff_loc} we still see a response when the pipe is approached at three different locations on the opposite side from the piezo elements, though the precise shape of the signal varies with approach location.

\begin{figure}[tb]
\centering
\begin{subfigure}[t]{0.45\linewidth}
\centering
{\includegraphics[width=1.1\textwidth]{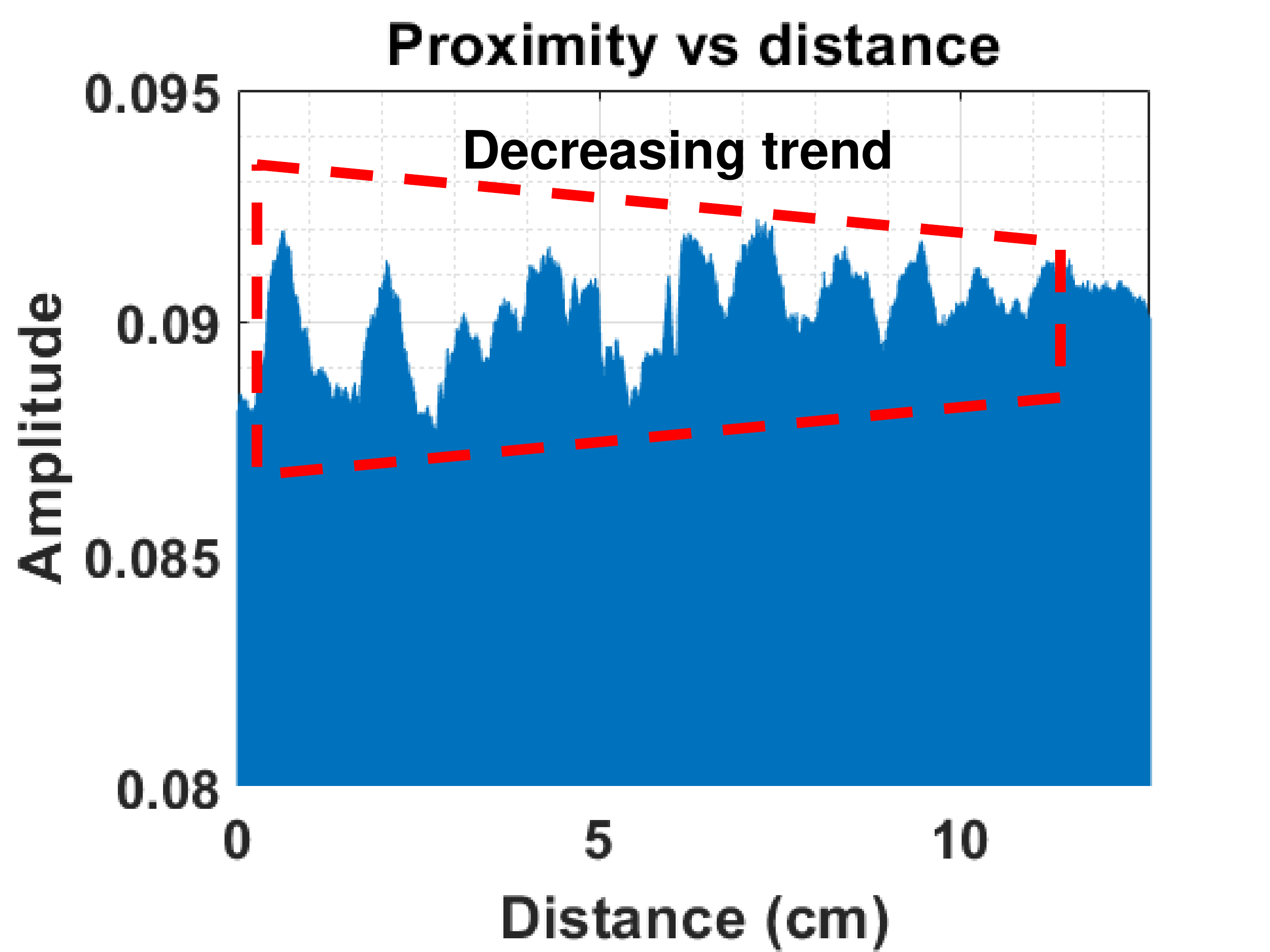}}
\caption{} 
\label{fig:amp_d}
\end{subfigure}
\begin{subfigure}[t]{0.45\linewidth}
\centering
{\includegraphics[width=1.1\textwidth]{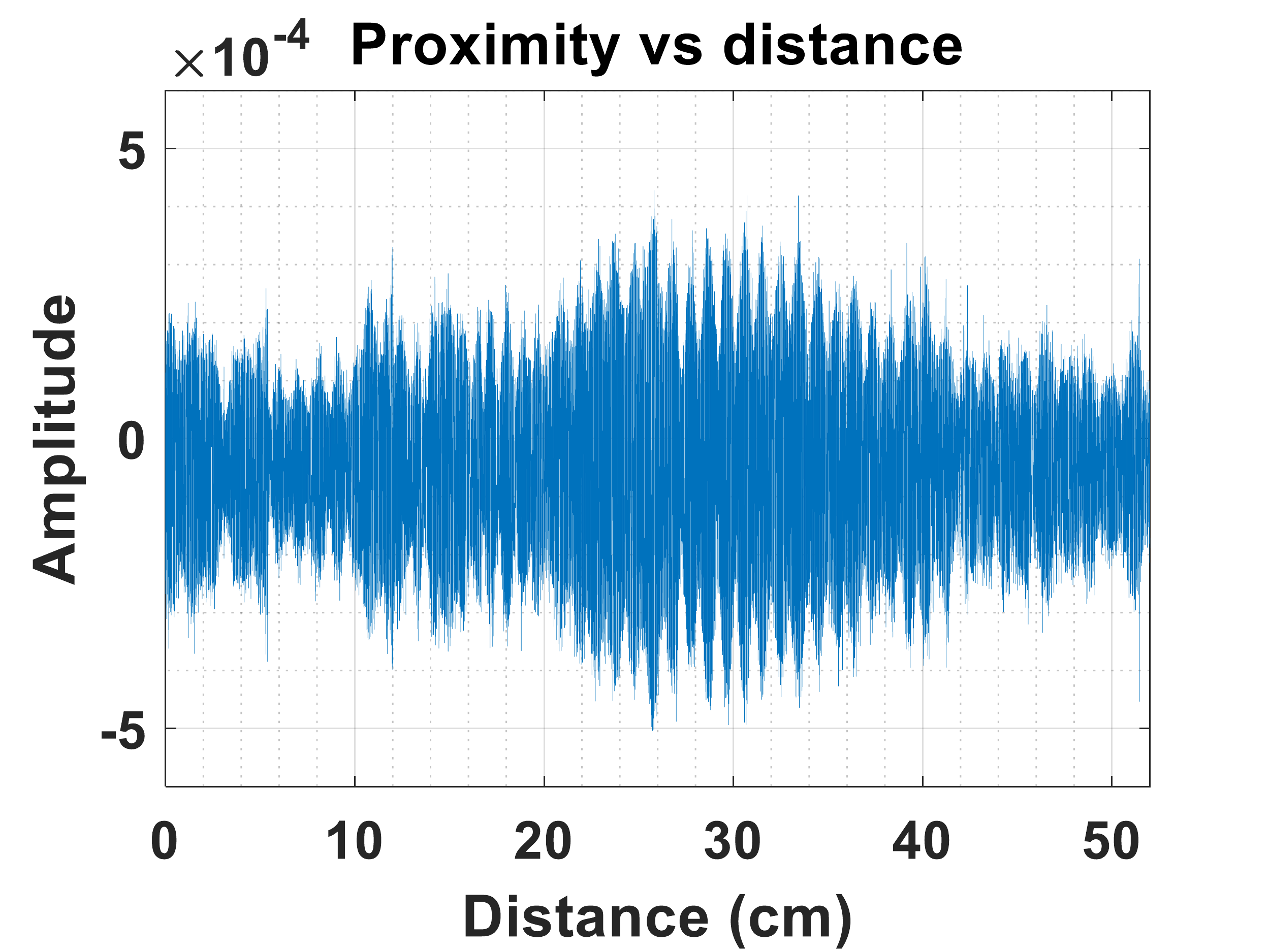}}
\caption{}
\label{fig:air}
\end{subfigure}
\caption{Recordings showing how the LSW signal is affected by the distance between the pipe and a wood block moving at constant velocity away from the pipe. (a) As the rod moves further away from the pipe, we see a decreasing amplitude pattern when the transmitter is attached to the pipe. (b) We do not observe this pattern when the transmitter is instead suspended 1 cm above the pipe. }
\vspace{-0.05in}
\label{fig:distance}
\end{figure}

Lastly, we designed a set of experiments to study how the proximity signal changes at different distances to the surface. In these experiments, we calculate the distance to the surface of the pipe by dragging a wooden stick about 4.5 cm in diameter using a motor, which moves the rod away from the surface of the aluminum pipe at a constant speed, v=0.49 cm/s. 
We show the results in Figure~\ref{fig:amp_d}. The proximity signal is distinct only when the rod is less than 10 cm from the pipe, and decreases in amplitude as the distance increases. In addition, the spacing between peaks is roughly 0.85 cm, which is around half the wave length ($\frac{\lambda}{2}=0.85$ cm at 20 kHz). This means the signal pattern is a function of the {\it distance} between the obstacle and the vibrating surface. As a control, we also measured the received signal when the piezo transmitter is detached from the surface (suspending it on a string about 1 cm above the surface of the pipe). We repeated the experiment with this setup, and show the results in Figure~\ref{fig:air}. We observe no decreasing signal pattern, and the signal instead appears random and uncorrelated with distance. We also note that the amplitude of the received signal is around 60 times smaller, as the signal no longer travels through the pipe surface and instead is over-the-air.
\vspace{-0.06in}

\subsection{On-robot LSW}
\label{subsec:lsw_robot}
\begin{figure}[tb]
\centering
\begin{subfigure}[t]{0.45\linewidth}
\centering
{\includegraphics[width=1\textwidth]{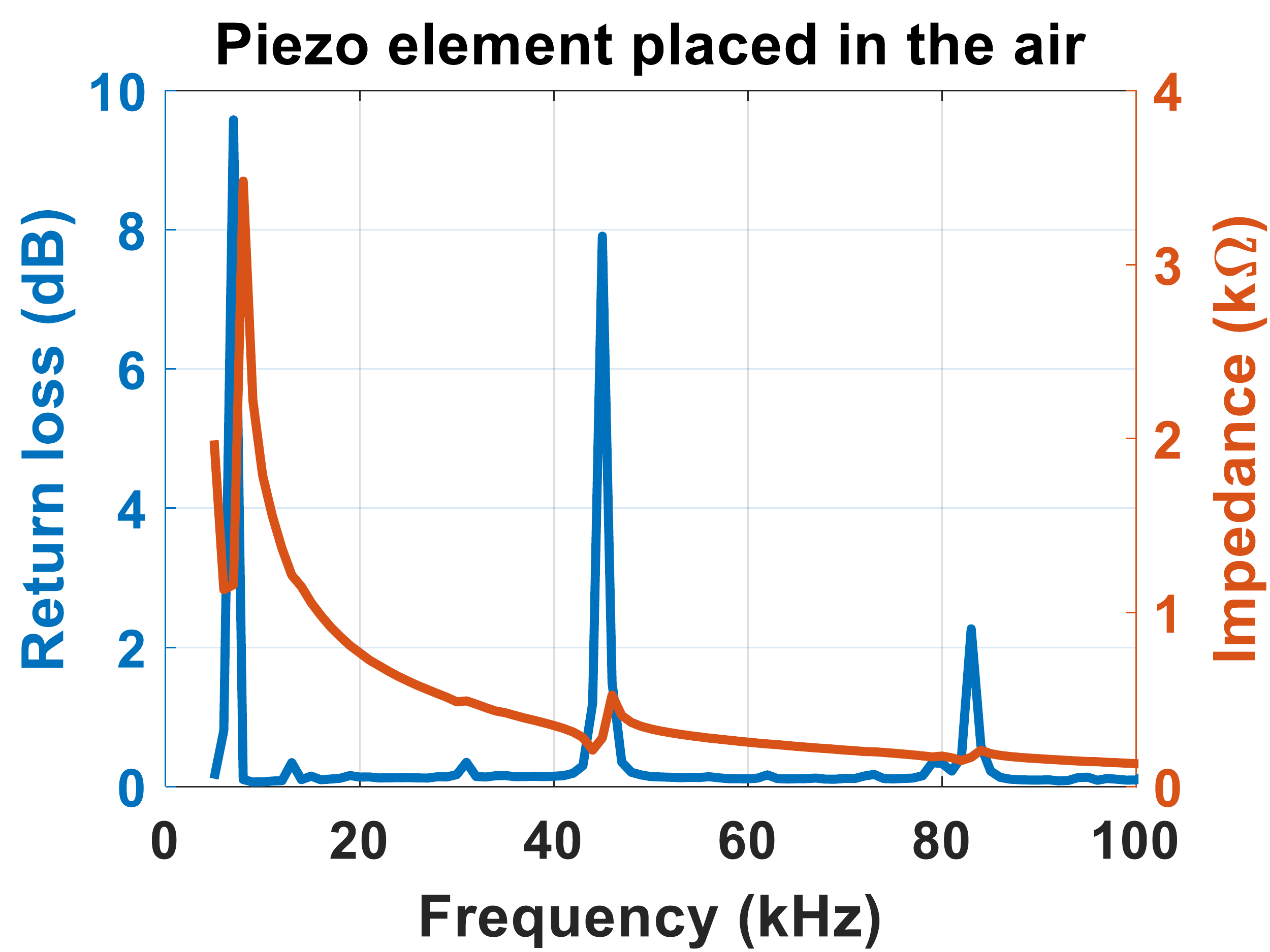}}
\caption{} 
\label{fig:air_couple}
\end{subfigure}
\begin{subfigure}[t]{0.45\linewidth}
\centering
{\includegraphics[width=1\textwidth]{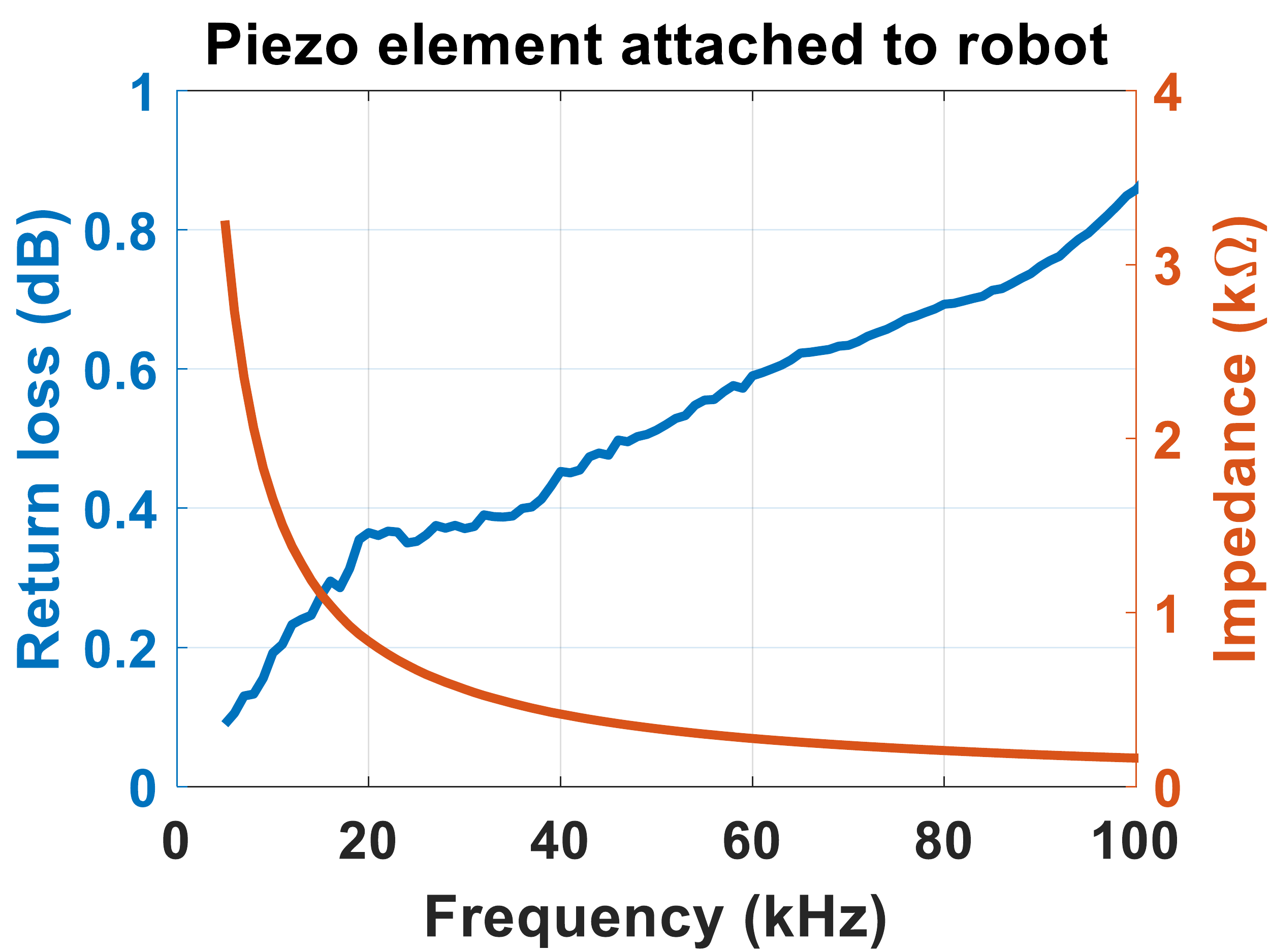}}
\caption{}
\label{fig:bot_couple}
\end{subfigure}
\caption{The return loss and impedance of a piezo element when (a) suspended in the air and (b) attached to the robot as described in Section \ref{subsec:lsw_robot}. We can see that the resonant frequency of the piezo element has changed as a result of robot mounting and that the Q-factor is dramatically reduced, making the transducer useful over a wide frequency range.}
\vspace{-0.2in}
\label{fig:couple}
\end{figure}

\begin{figure}[tb]
\centering
\begin{subfigure}[t]{0.44\linewidth}
\centering
{\includegraphics[width=0.9\textwidth]{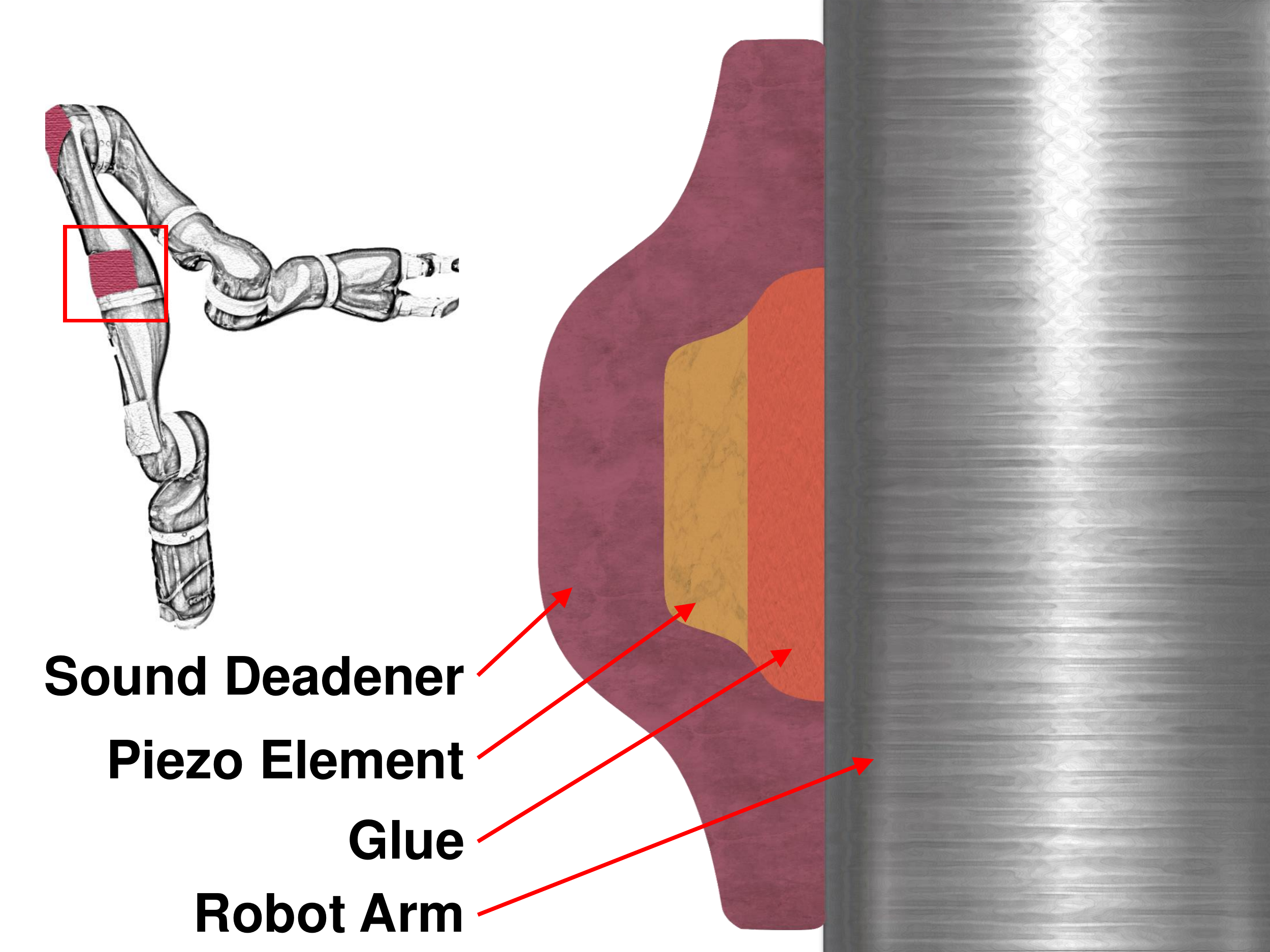}}
\caption{The mounting detail for the piezo element. } 
\label{fig:illustration_couple}
\end{subfigure}
\begin{subfigure}[t]{0.45\linewidth}
\centering
{\includegraphics[width=1\textwidth]{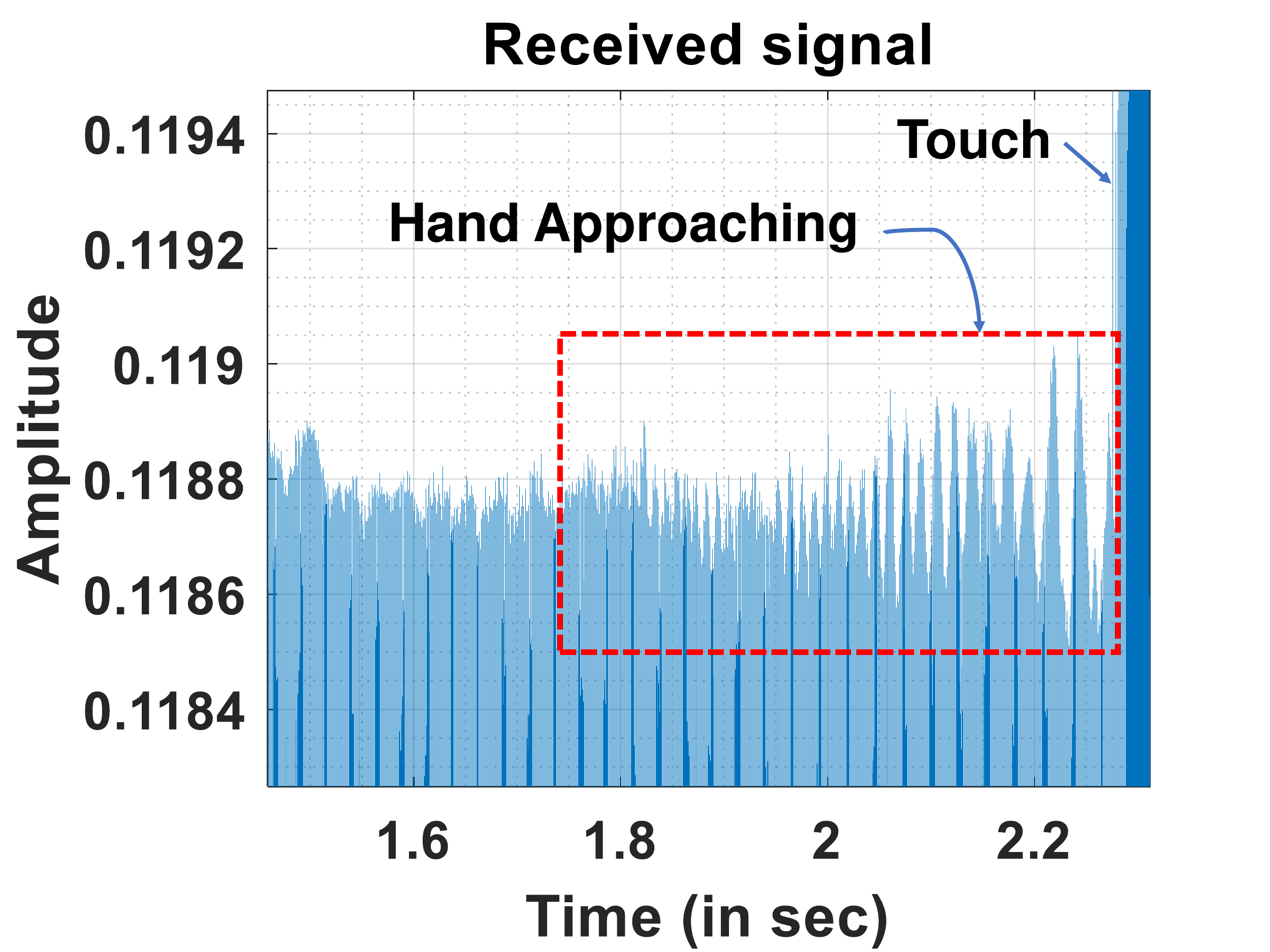}}
\caption{The proximity signal measured on a Kinova Jaco manipulator.}
\label{fig:robot_lsw}
\end{subfigure}
\caption{Our proposed method for coupling the piezo transducers with a robot. (a) A schematic of our mounting system. The piezo element couples with the robot, and does not leak meaningful signal to the air. (b) A proximity detection test performed with the system mounted on a robot, similar to that in Figure \ref{fig:sensing_demo_zoom}. The proximity signal can be observed with robot mounted transducers.}
\vspace{-0.2in}
\label{fig:demo_robot}
\end{figure}

While the previous section established the useful properties of the LSW, this section describes how we use that signal to design a proximity detection system for use with an off-the-shelf robot arm. We implement \systemname on a Kinova Jaco~\cite{jaco} arm, the surface of which is composed of carbon fiber, 
a material that is far from ideal as a vibrating surface, but which demonstrates how the LSW can be used outside ideal conditions. 
Key to implementing LSW sensing on this robot is to ensure the piezo elements couple with the robot surface- Piezo elements are usually designed to couple with air. In Figure~\ref{fig:air_couple} we show how a piezo element coupled with air resonates at 7 kHz, 45 kHz, and 83 kHz, and most of the energy transfers into the air. \footnote{We measured the characteristics of our piezo element with an Array Solutions VNA UHF vector network analyzer~\cite{vna}.} In order to decouple the piezo element from the air, we used a thermoplastic adhesive (glue) to attach the piezo element to the surface of the robot, and sealed the piezo element with a layer of Noico solutions 80 mil sound deadener~\cite{dead}, as illustrated in Figure~\ref{fig:illustration_couple}. After attaching the piezo element to the robot this way, we performed the same measurement and show the results in Figure~\ref{fig:bot_couple}. We observed that the resonances that pumped acoustic energy into the air were removed, and the impedance indicated the piezo element became almost purely capacitive. We tested this robot instrumentation in a manner similar to the experiments in Section \ref{subsec:phy} and show the results in Figure~\ref{fig:robot_lsw}. We observed a similar signal from the robot sensors to that from the pipe, which shows we can successfully use the LSW to detect proximity on a commercial robot.

In \systemname, the LSW is transmitted from one of the piezo elements. It travels through the robot surface and is received by another piezo element. The robot arm functions as the wireless channel~\cite{erceg1999empirically} in this system. Denoting the transmitted signal as $s(t)$ and the wireless channel as $h$. The received signal $r(t)$ can be represented as:
\begin{equation}
	\begin{aligned}
      &r(t) = h \times s(t)+\nu.
	\end{aligned}
	\label{eqn:channel}
\end{equation}
Where $\nu$ is the internal and external system noise. Our high level goal is to detect if there is an approaching object from $r(t)$ under the time varying channel response $h$ and noise $\nu$.

\begin{figure}
\centering
\begin{minipage}{.485\linewidth}
  \centering
{\includegraphics[width=0.95\textwidth]{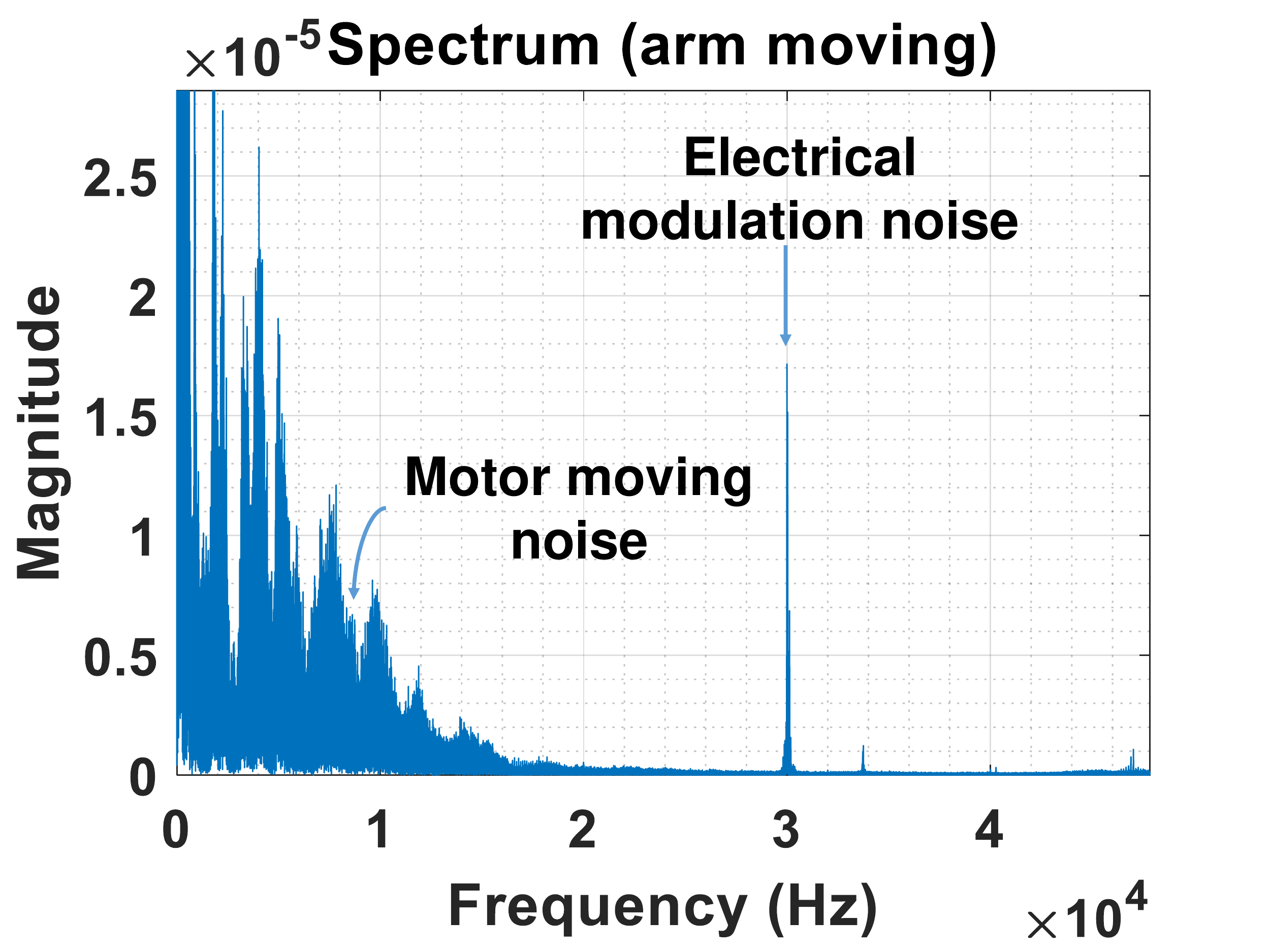}}
\caption{Noise spectrum extracted from a recording of the robot arm moving for 10 minutes, with no excitation signal applied to the piezo element. Common motor noise frequencies are less than 18 kHz, with electrical noise at 30kHz.}
\label{fig:noise}
\end{minipage}%
\hspace{0.1cm}
\begin{minipage}{.485\linewidth}
  \centering
\includegraphics[width=0.95\textwidth]{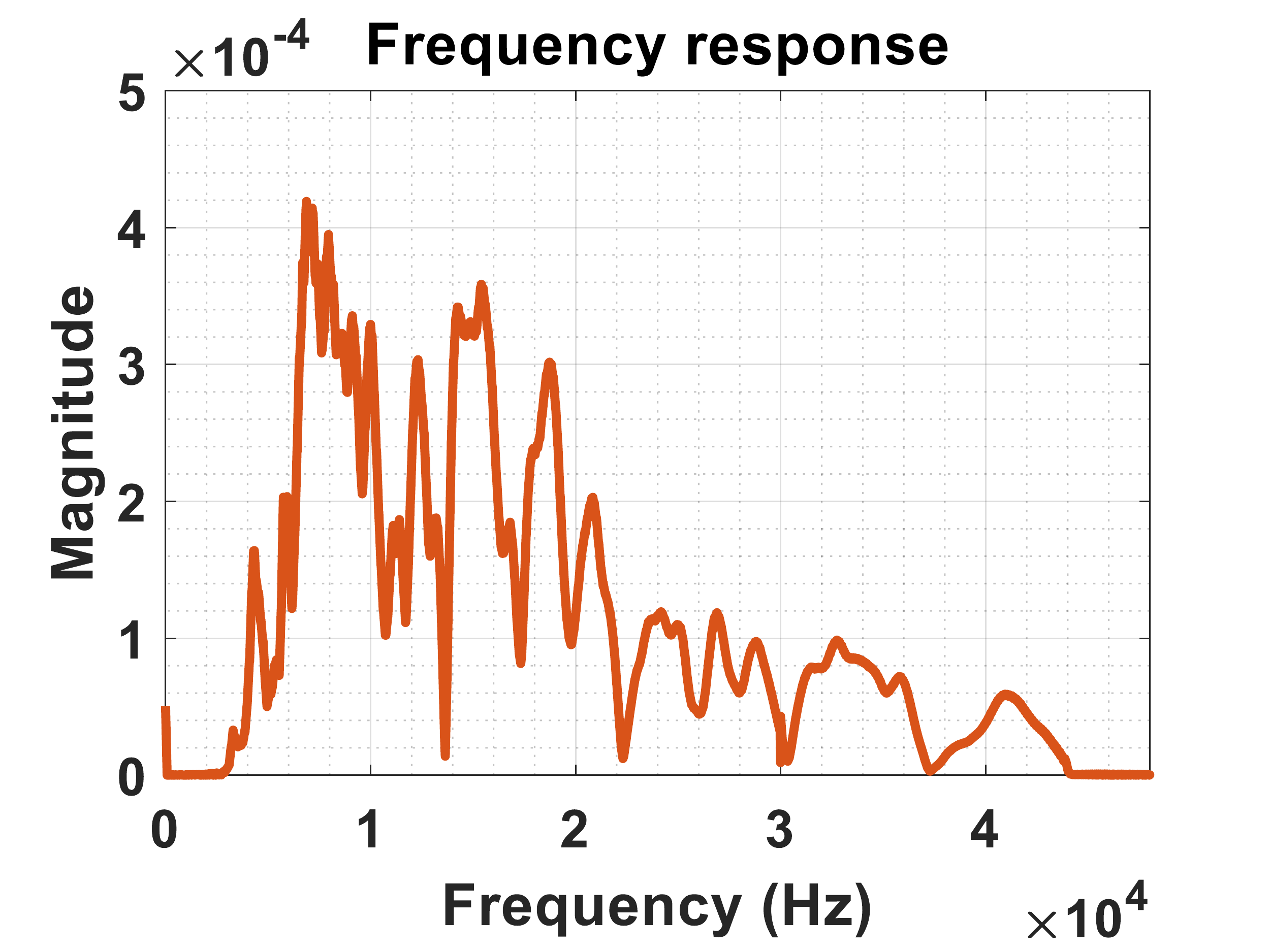}
\caption{The on robot frequency response from a recording of the piezo transmitter that sweeps from 0-48 kHz with equal intensity in each frequency bin. The 18-19 kHz range both produces good frequency responses and avoids most robot noise.}
\label{fig:fr}
\end{minipage}
\vspace{-0.15in}
\end{figure}

\vspace{4pt}\noindent\textbf{Noise Reduction:} Having observed the LSW effect on a robot, the next challenge to address is the noise $\nu$ from the robot arm. The robot introduces two type of noise: (1) Electrical noise from the power supply modulating the motors, both when the arm moves and when it is stationary, and (2) mechanical noise from the motors and gears when the arm is in motion. Figure~\ref{fig:noise} shows a noise spectrum recorded while the robot is moving and we set the pose of each joint independently by choosing a value uniformly at random. We observe that the majority of the mechanical noise energy resides in a sub-15 kHz range, while there is also a significant electrical noise spike at 30 kHz.

Once we understand the on-robot noise characteristics, we can choose a range of useful frequencies which don't overlap with the noise spectrum. However, due to the acoustically non-homogeneous nature of the robot arm~\cite{fan2020acoustic}, the LSW behaves differently in different frequency bins. Figure~\ref{fig:fr} shows the frequency response measured by the piezo receiver from the transmitter. From this figure, we surmise that the 18-19 kHz range is a sweet spot for both avoiding noise and sending energy efficiently through the robot. As such, in \systemname design, we choose a $f = 19$ kHz continuous sinusoidal signal $s(t)=\sin{2\pi f t}$ to excite the piezo transmitter, and apply a narrow 19 kHz bandpass filter to the received signal for a further denoising. 

\begin{figure}[tb]
\centering
\begin{subfigure}[t]{0.45\linewidth}
\centering
{\includegraphics[width=1\textwidth]{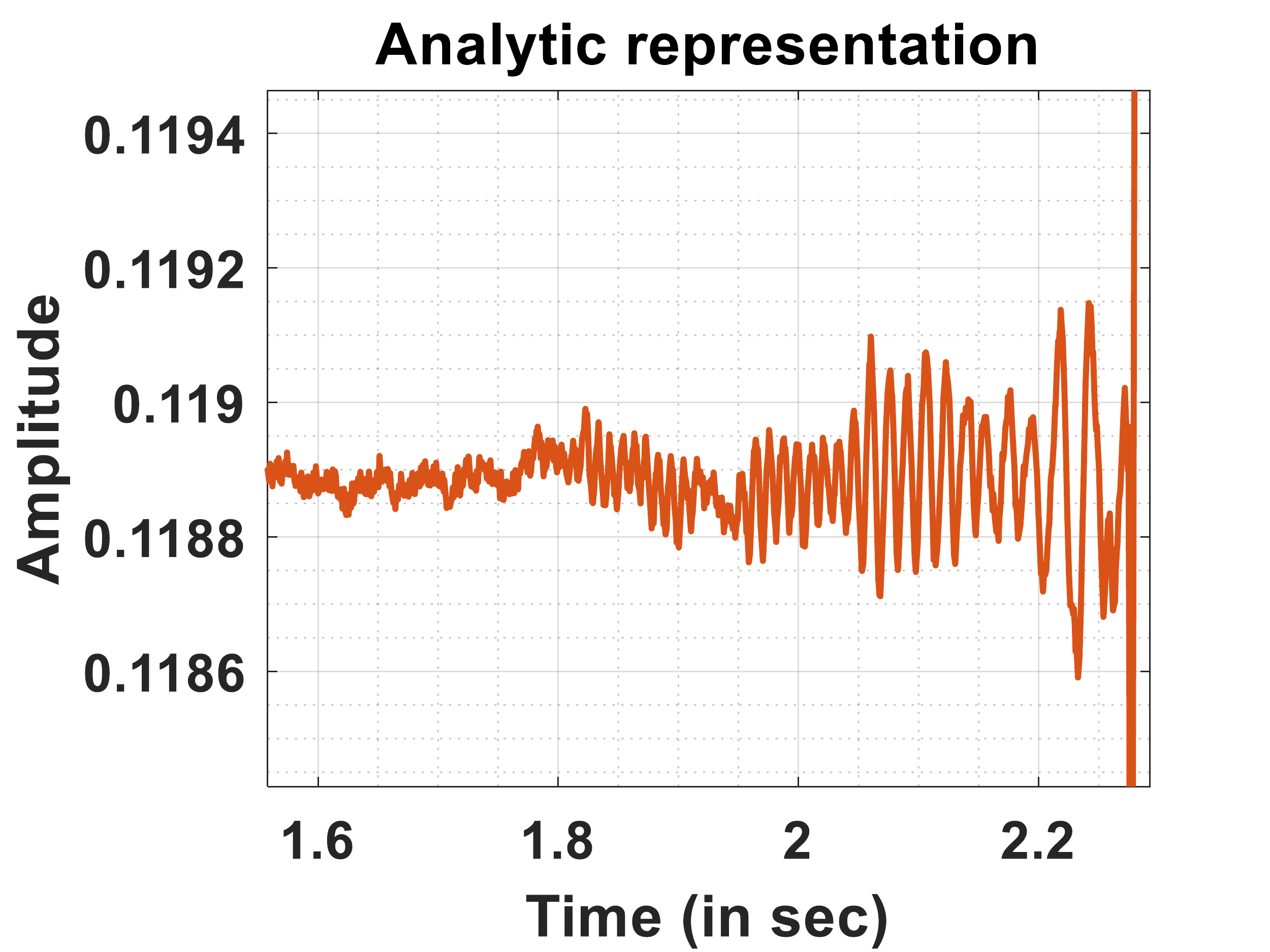}}
\caption{} 
\label{fig:analytic}
\end{subfigure}
\begin{subfigure}[t]{0.45\linewidth}
\centering
{\includegraphics[width=1\textwidth]{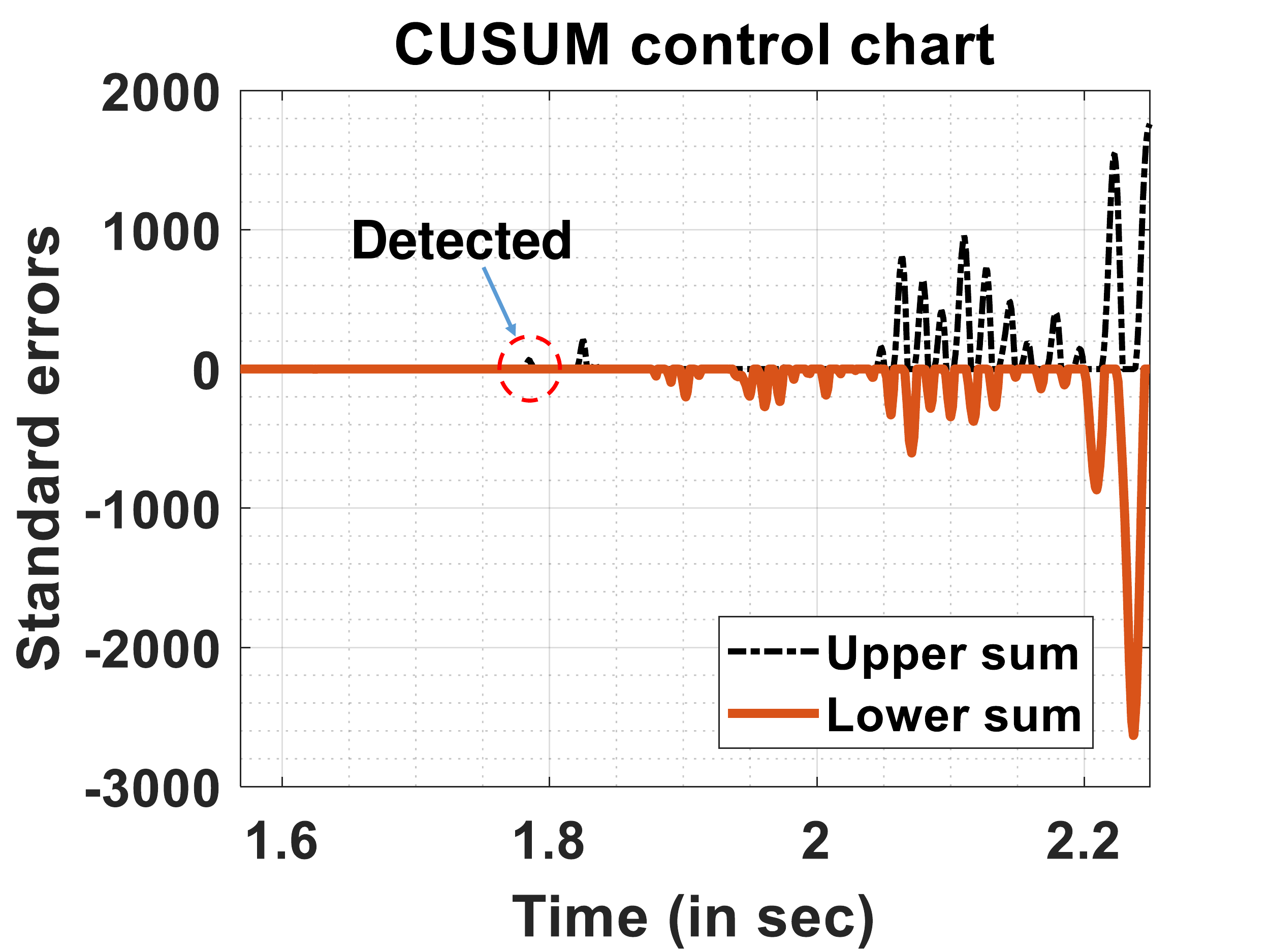}}
\caption{}
\label{fig:cusum}
\end{subfigure}
\caption{(a) Analytic representation of the proximity signal. (b) The CUSUM control chart for the analytic signal in (a). The proximity pattern is detected around 1.8 second in this example.}
\vspace{-0.15in}
\label{fig:disturb}
\end{figure}

\vspace{4pt}\noindent\textbf{Recovering the proximity signal:} Now that we have selected a transmitter frequency and removed robot noise, we must recover the proximity signal from the LSW. This signal is orders of magnitude weaker than the surface guided signal, as evidenced by Figure~\ref{fig:signal_demo} and Figure~\ref{fig:robot_lsw}. The large through-surface signal hinders the detection process and makes it difficult to use machine learning models to account for other factors influencing the signal, such as self-proximity, as the relevant differences are a small fraction of the overall magnitude. To account for this, after the received signal $r(t)$ has been filtered by the band pass filter, we calculate its analytic representation. Denoting $x(t)$ as the {\it filtered} received signal, the analytic signal $x_{a}(t)$ can be represented as: 
\begin{equation}
	\begin{aligned}
      &x_{a}(t) \triangleq x(t)+jy(t).
	\end{aligned}
\end{equation}
Where $j$ is an imaginary unit. $y(t)$ is the Hilbert transform of $x(t)$, which is the convolution of $x(t)$ with the Hilbert transform kernel $h(t) \triangleq \frac{1}{\pi t}$.
As a result, the analytic representation is a linear time-invariant filter process that can be written as: 
\begin{equation}
	\begin{aligned}
      &x_{a}(t) = \frac{1}{\pi}\int_{0}^{\infty}X(w)e^{jwt} dw.
	\end{aligned}
\end{equation}
Where $X(w)$ is the Fourier transform of the signal $x(t)$. In our real time implementation, we apply this process every $L$ samples. Figure~\ref{fig:analytic} shows the analytic representation of the proximity signal from Figure~\ref{fig:robot_lsw} with $L=300$ (sampling rate is 96 kHz). The proximity pattern is represented more obviously in the analytic signal. We employ a statistical quality control algorithm, CUSUM~\cite{granjon2013cusum}, to monitor the received signal analytic. 
Specifically, CUSUM 
detects if there is a significant change or disturbance in a time series sequence. As shown in Figure~\ref{fig:cusum}, the proximity pattern is detected by the CUSUM algorithm around 1.8 second.

\begin{figure}[t]
\centering
\begin{subfigure}[t]{0.45\linewidth}
\centering
{\includegraphics[width=1\textwidth]{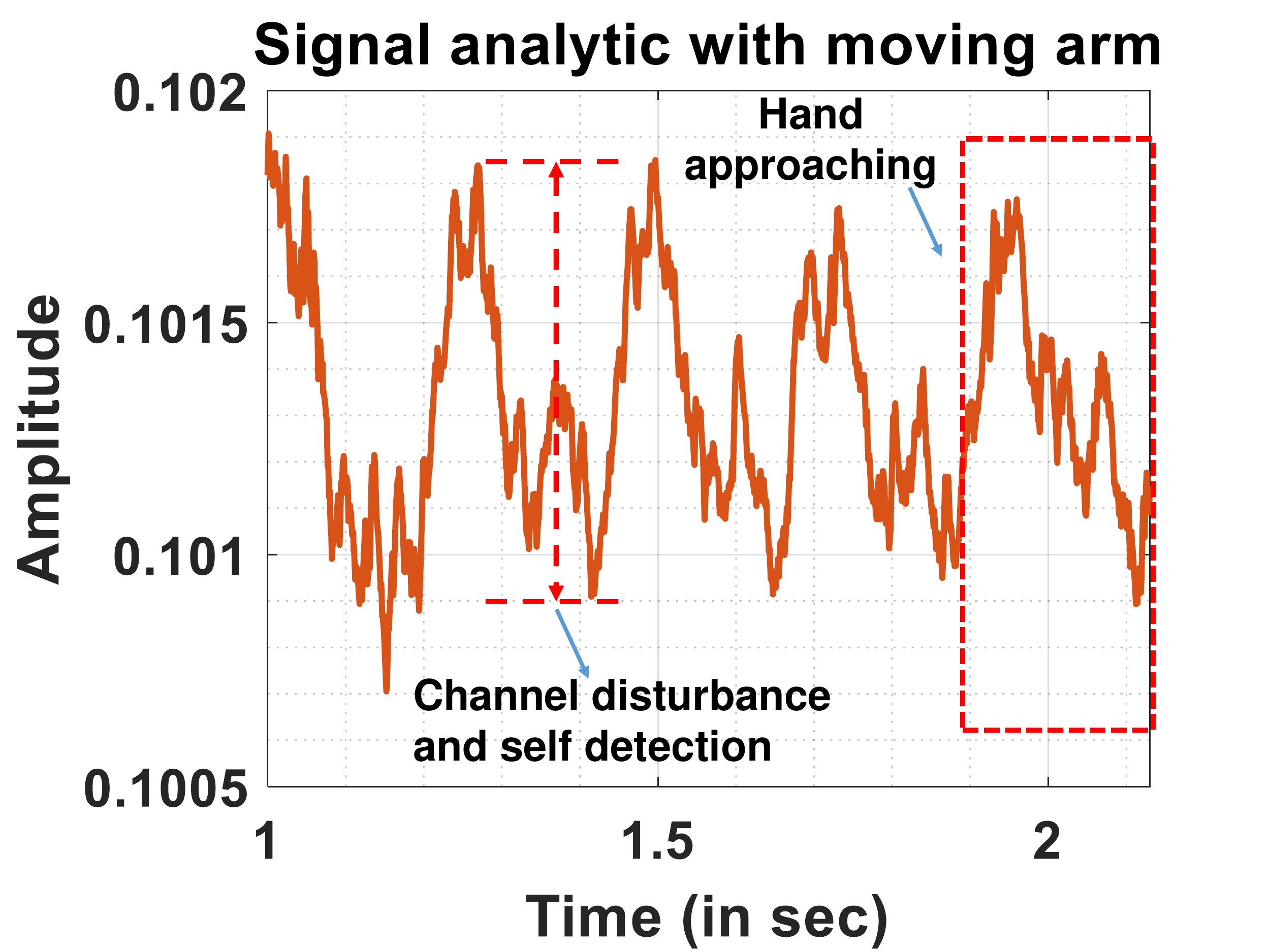}}
\caption{} 
\label{fig:mobile_bot}
\end{subfigure}
\begin{subfigure}[t]{0.45\linewidth}
\centering
{\includegraphics[width=1\textwidth]{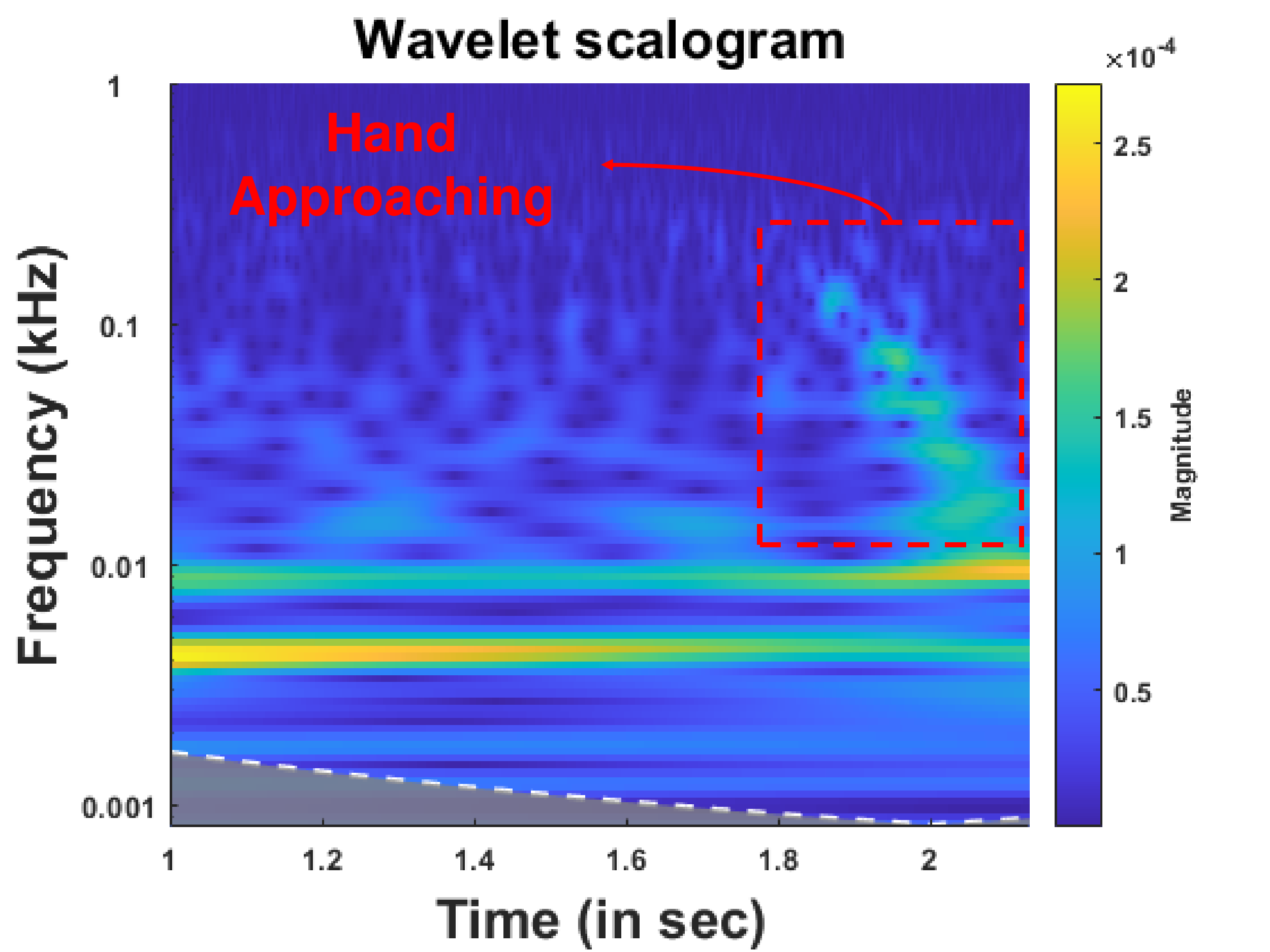}}
\caption{}
\label{fig:wavelet}
\end{subfigure}
\caption{(a) The proximity pattern from a hand approaching a moving robot (at 1.8-2.2 seconds) is hard to see from the signal analytic, (b) but the pattern is easily visible in the scalogram.}
\vspace{-0.2in}
\end{figure}

\vspace{4pt}\noindent\textbf{Channel disturbance and self detection:}
While the analytic proximity signal is easier to detect with this processing, a number of challenges for discriminating real signals from false positives remain. The wireless channel $h$ in Equation~\ref{eqn:channel} is mainly determined by the mechanical characteristics of the robot- for example, the pose of the robot and the internal gear arrangement. Therefore, when the arm moves it changes $h$ and alters the received signal $r(t)$ as well as its analytic representation $x_{a}(t)$. Additionally, the robot arm could detect itself as an “obstacle” as linkages move closer to each other, meaning not all proximity signals should be reacted to. Figure~\ref{fig:mobile_bot} shows a snapshot of the signal analytic when the arm is moving. As illustrated in the figure, there is a strong signal variation caused by the channel disturbance and self detection. In this recording a person's hand approaches at 1.8-2.2 seconds, but the proximity pattern is hard to distinguish visually. However, the proximity information is not lost-- Figure~\ref{fig:wavelet} shows the scalogram of this signal analytic using wavelet analysis. We see a strong channel disturbance and a self-detection signal, but the proximity signal is also visible at 1.8-2.2 seconds.
\vspace{-0.15cm}

\subsection{Classification}
\label{subsec:classification}

While a human can discriminate visually between approaching objects and background noise once familiar with the data, automatic detection is non-trivial. The factors discussed above can have highly nonlinear effects on the signal, 
which make discriminating signals typical of approaching obstacles difficult. The complexity of this task motivates a machine learning approach to the problem.



\vspace{4pt}\noindent\textbf{Learning proximity:} To address these issues, we use a 1D convolutional neural network to perform binary classification on windowed analytic signals, classifying each segment as corresponding to an approaching obstacle or to the negative case where no obstacles are close by. Our network uses a simple fully-convolutional architecture, consisting of 7 1D convolutions, plus a linear layer outputting a scalar prediction between 1 (object approaching) and 0 (no object approaching). This classifier takes as input windows of 960 samples collected at a 96 kHz sample rate, such that each window corresponds to 0.01 seconds of audio. The samples are normalized in 0.3 second windows (our detector window size below) before being input to the network.
Hyperparameters and the training process are discussed in Section \ref{subsec:network_training}.

\vspace{4pt}\noindent\textbf{Detector:} While this classifier achieves high accuracy on 0.01 second timescales, for robust detection at human timescales aggregation across multiple 0.01 second windows is needed. To make a final decision, we pass $N$ sequential $0.01$-second window scalar predictions into a larger window detector, average them, and make the final determination on whether an approaching obstacle has been detected or not by thresholding the average. During our on-robot testing we found the classifier's predictions are not independent and identically distributed among  0.01 second windows, so the detector uses $N=30$, a 0.3 second sliding window, to compensate. We slide the window in increments of 0.1 seconds. When the detector's threshold is exceeded, a stop signal is sent to the robot, which freezes in place to avoid or minimize the impact. 

\vspace{-0.05in}

\section{Implementation}
\label{s:impl}
We describe the implementation details in this section. 
\vspace{-0.05in}
\subsection{Hardware Setup}
For \systemname, we deploy two CPT-2065-L100 piezo elements~\cite{piezo} 20 cm apart on a Kinova Jaco Gen 2 7-DoF manipulator. We use a SMSL M100 digital-to-analog converter (DAC)~\cite{M100} to decode the digital waveform from the computer. We pass this converted analog signal to a Cavalli Liquid Carbon X amplifier~\cite{carbon}, which directly drives the transmitting piezo element. The receiving piezo element is connected to a Zoom F8n MultiTrack Field Recorder~\cite{recorder}. This field recorder is grounded on the robot's ground to eliminate ground loop capacitive coupling. We use a sampling rate $Fs$ of 96 kHz. Signal processing is done on an Intel NUC7i7BNH computer~\cite{nuc}. Both piezo elements are attached to the surface of the robot with thermoplastic adhesive and covered by a sound deadening material as described in Section~\ref{subsec:lsw_robot}.

\vspace{-0.05in}

\subsection{Neural Network Training}
\label{subsec:network_training}
To train our classification network, we used the following implementation details and hyperparameters. All neural networks were implemented in PyTorch 1.7.1~\cite{pytorch}.
Our network consists of 7 1D convolution layers with length 7 kernels and 256 hidden channels, with a stride of 2 at each layer. We trained this network using softmax cross-entropy with the Adam optimizer \cite{DBLP:journals/corr/KingmaB14}. We used ReLU non-linearities, and batch normalization \cite{ioffe2015batch} with a batch size of 32 samples, with 16 positive and 16 negative samples per batch. Networks were trained 
at a learning rate of 0.00001.
To sample training batches, 32 new random windows were selected from among all possible contiguous 960-sample windows in the positive and negative training sequences. 
Computing predictions for a 0.3 second window using our implementation takes about 0.0025 seconds on a 3090 GPU in a high-end workstation.


\vspace{-0.0in}

\section{Evaluation}
\label{s:evaluation}
\vspace{-0.02in}
We present our experiments and evaluation results in this section. We first describe a set of micro-benchmark experiments to deepen our understanding of LSW signals, followed by realistic on-robot experiments to demonstrate how the LSW can be used for practical proximity detection.
\vspace{-0.1in}
\begin{figure*}[t]
\begin{minipage}[t]{0.24\textwidth}
     \centering
{\includegraphics[width=1\linewidth]{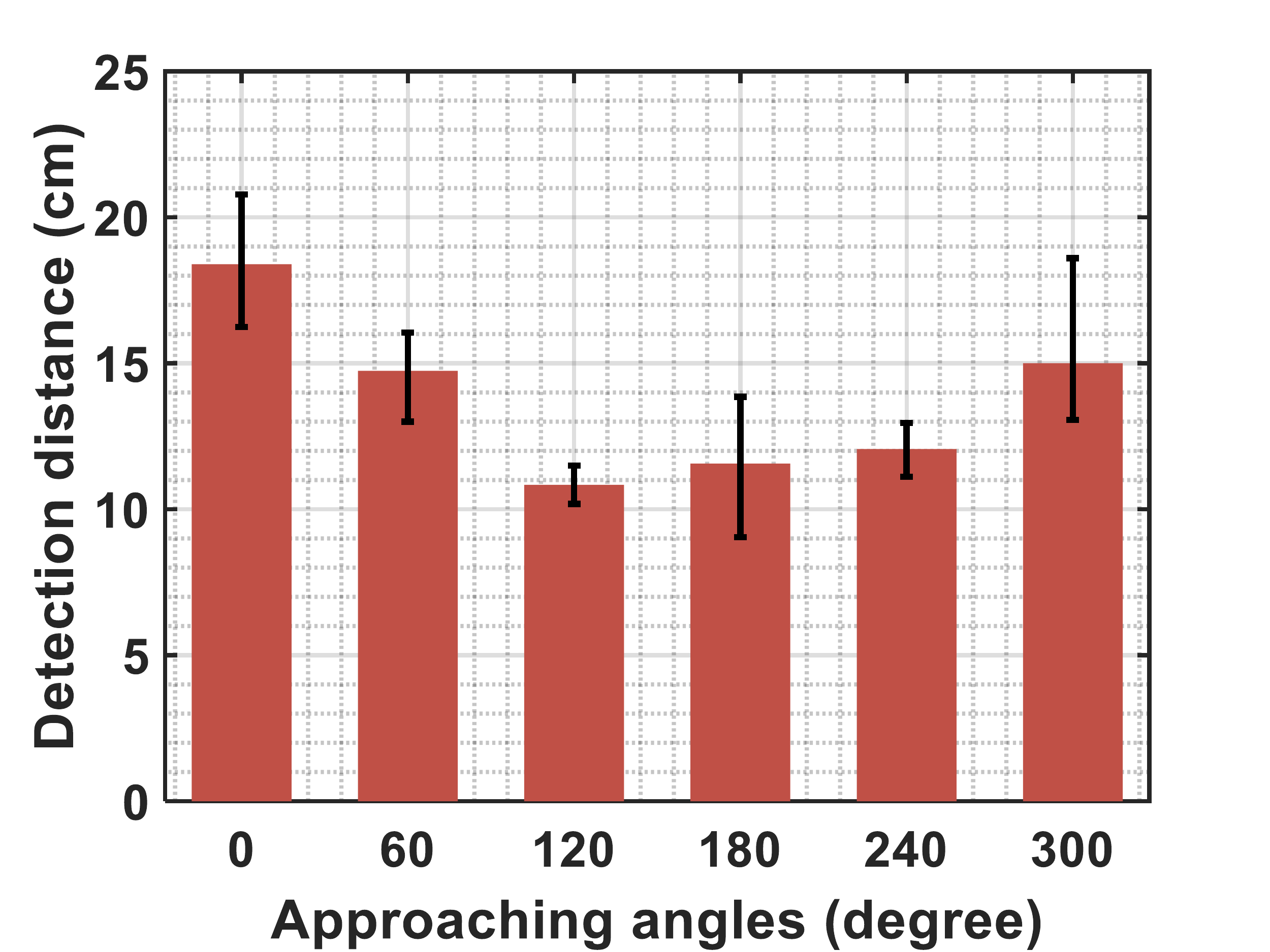}}
\caption{Max detection distance vs. approaching angle.}
\label{fig:d_a}
    \end{minipage}\hfill
\begin{minipage}[t]{0.24\textwidth}
     \centering
{\includegraphics[width=1\linewidth]{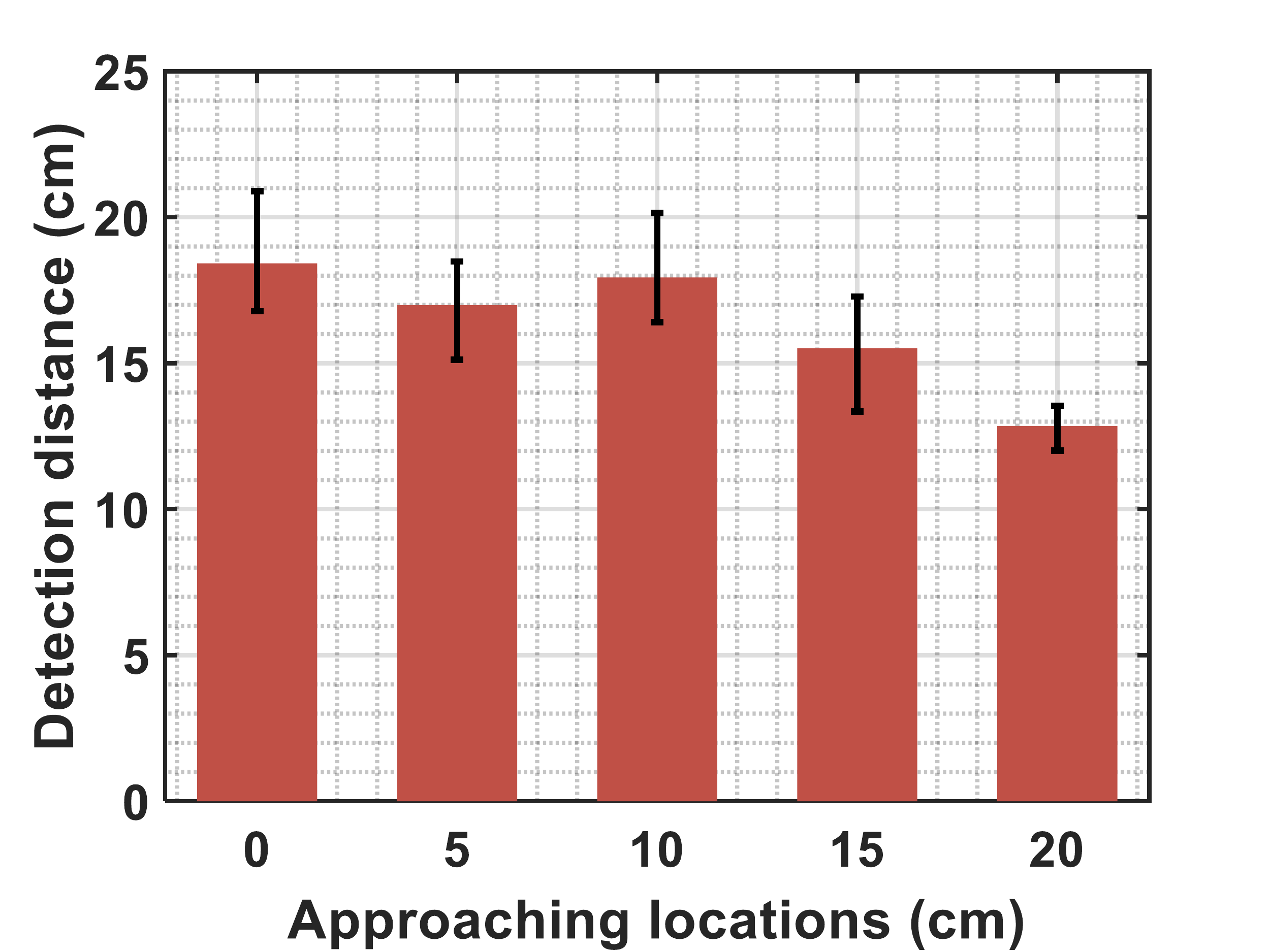}}
\caption{Max detection distance vs. approaching location.}
\label{fig:d_l}
\end{minipage}\hfill
    \begin{minipage}[t]{0.23\textwidth}
     \centering
{\includegraphics[width=0.95\linewidth]{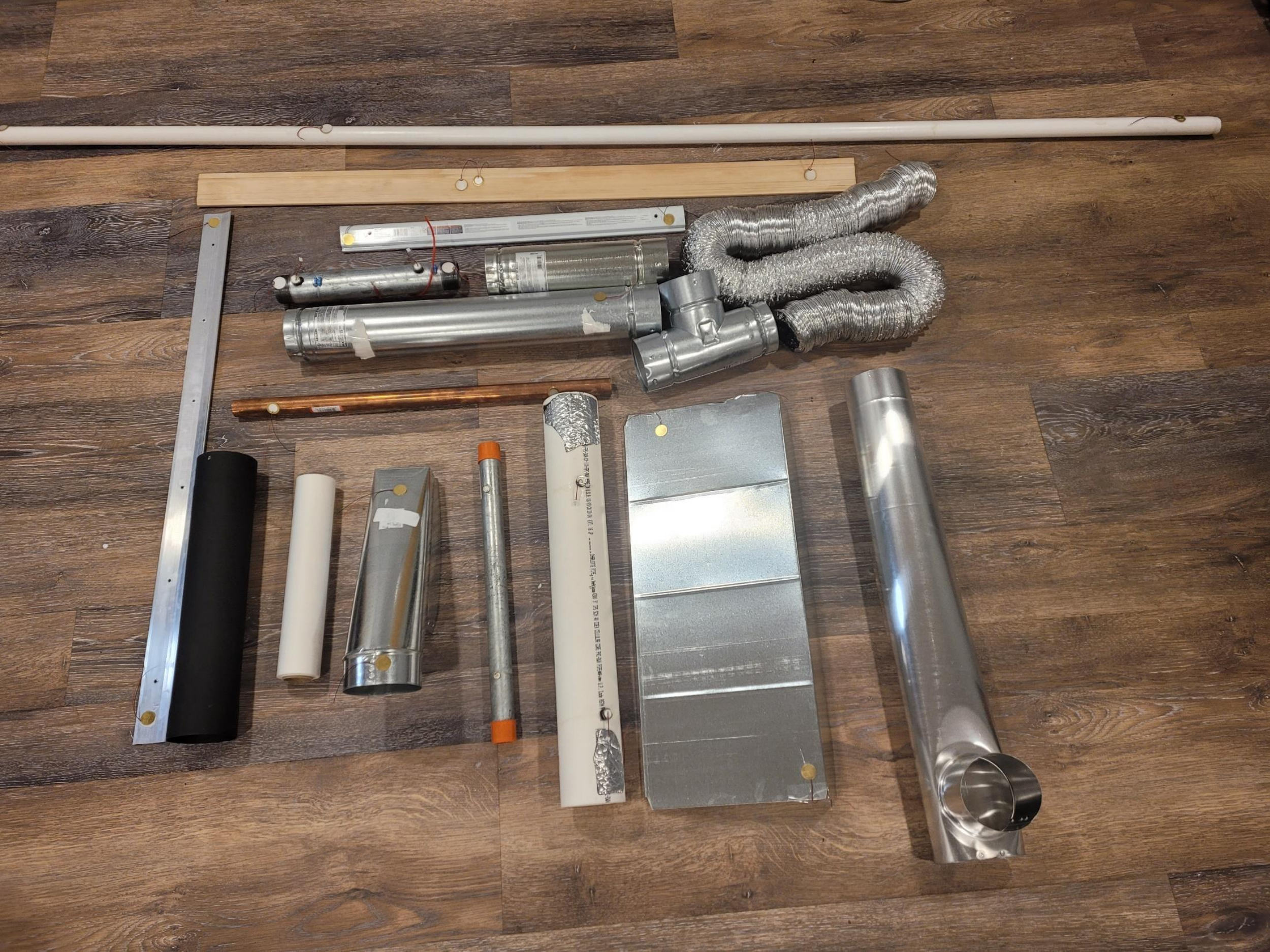}}
\caption{Materials we used as ``robots'' .}
\label{fig:dummy}
   \end{minipage}\hfill
     \begin{minipage}[t]{0.24\textwidth}
     \centering
{\includegraphics[width=1\columnwidth]{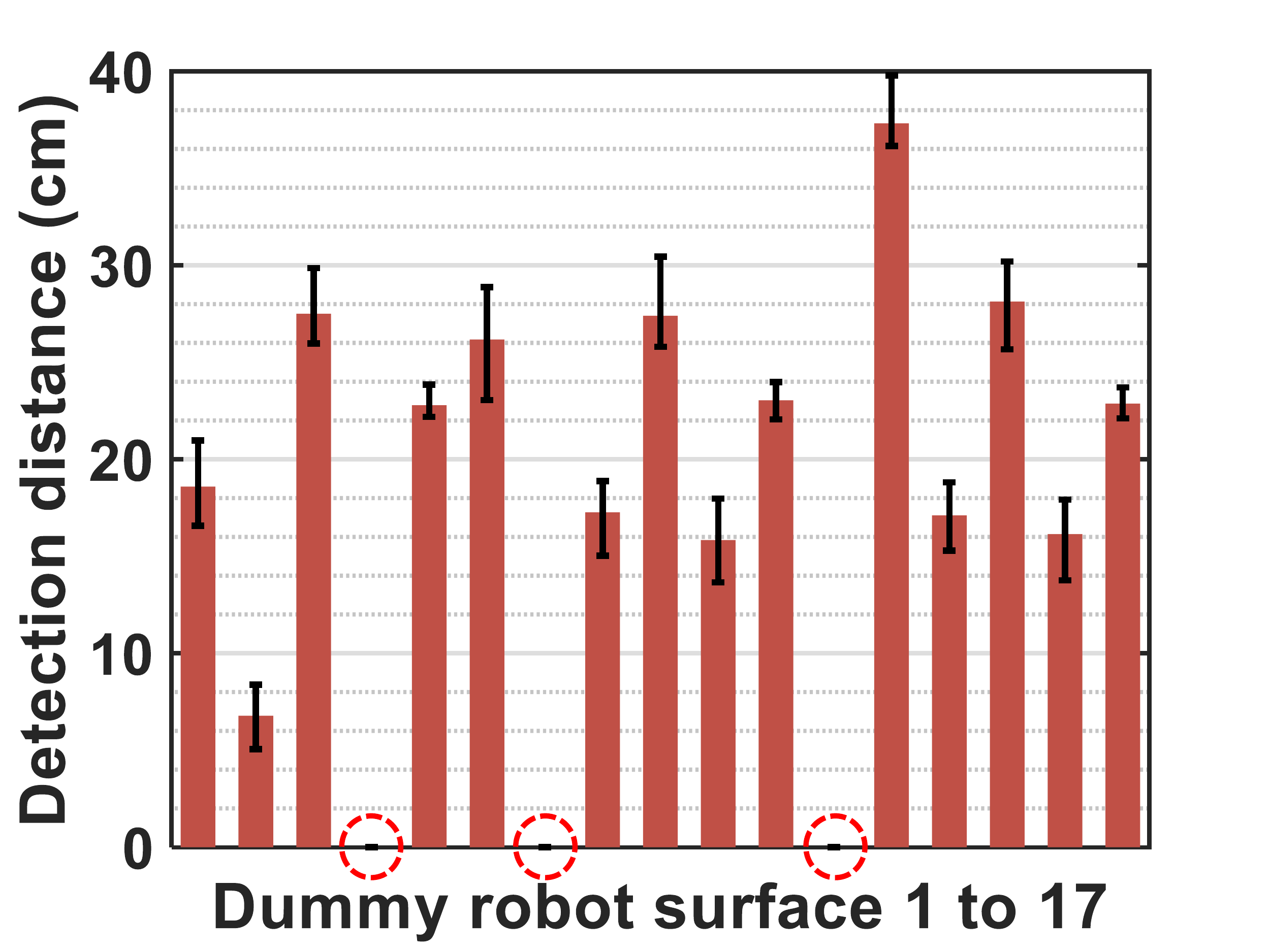}}
\caption{Max detection distance vs. ``robot'' surfaces.}
\label{fig:d_o}
   \end{minipage}\hfill
\vspace{-0.2in}
\end{figure*}

\vspace{-0.05in}

\subsection{Micro-benchmark}
\label{subsec:microbench}
In this section, we study the fundamental properties of the LSW. We use the motor controlled stick mentioned in Section~\ref{subsec:phy} to approach stationary targets instrumented with piezo elements. The stick approaches the target surface at a constant speed until the rod hits the surface, allowing us to know the zero distance time point. We use the CUSUM algorithm (introduced in Section~\ref{subsec:lsw_robot}) to detect (1) if there is a significant change in the signal analytic caused by the stick approaching, and (2) how far away from the target surface we can detect the approaching stick.

\begin{figure}
\centering
\begin{minipage}{.465\linewidth}
  \centering
{\includegraphics[width=0.95\textwidth]{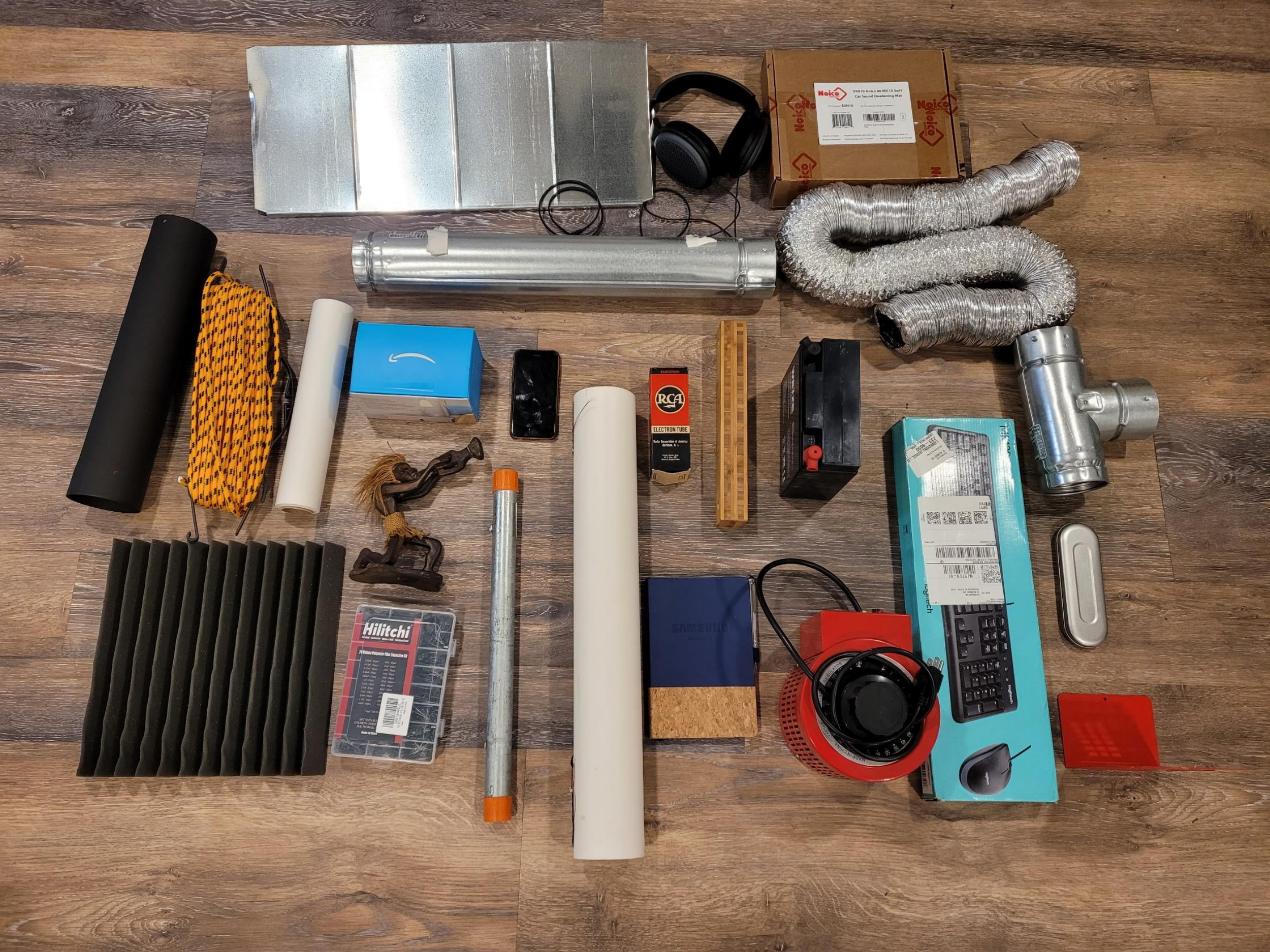}}
\caption{Objects used in the field study.}
\label{fig:ob}
\end{minipage}%
\hspace{0.1cm}
\begin{minipage}{.485\linewidth}
  \centering
\includegraphics[width=0.95\textwidth]{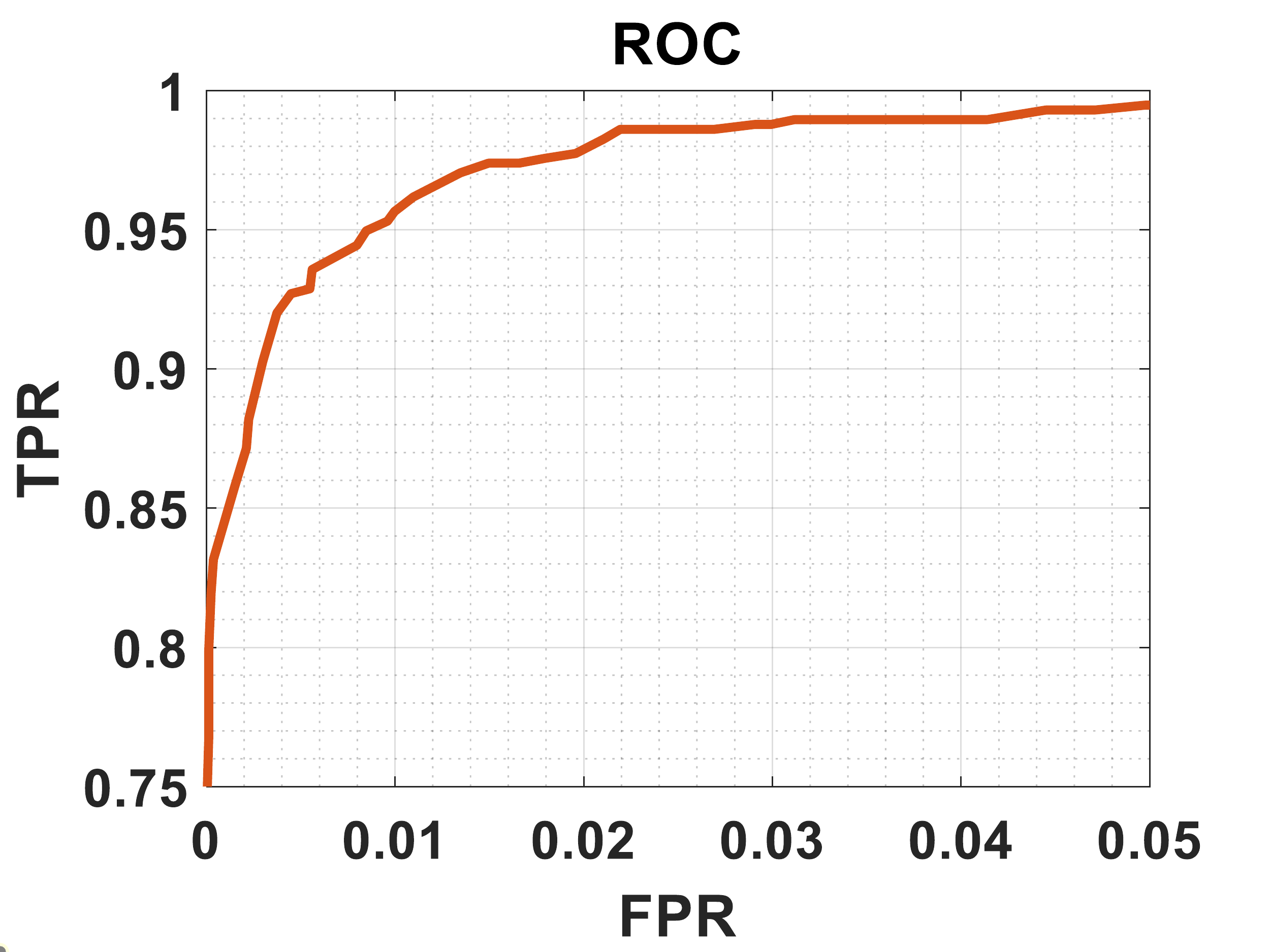}
\caption{The ROC curve for our detector.} 
\label{fig:roc}
\end{minipage}
\vspace{-0.275in}
\end{figure}

\subsubsection{Detecting proximity vs. approach angle}
In our first experiment, we approached the midpoint between the piezo transmitter and receiver on the robot arm from different angles using our stick apparatus. This tests the omnidirectional sensing capabilities of \systemname. We denote 0 degrees as the orientation perpendicular to the plane of the transmitter and receiver. 
We sample angles of 60, 120, 180, 240, and 300 degrees offset from that angle to test approaches from all directions.
We performed this experiment 13 times for each angle. Figure~\ref{fig:d_a} shows the results. They indicate that \systemname can detect the approaching stick from all angles tested. Interestingly, the maximum detection distance is shorter when approaching the side opposite the piezo elements. Among angles tested, the shortest maximum detection range was 10.9 cm at 120 degrees, an oblique angle on the opposite side of the robot from the piezo elements, compared to a maximum detection range of 18.3 cm when approaching perpendicular to the piezo elements on the same side.

\subsubsection{Detecting proximity vs. approach location} Next, we explore the relationship between the position of the stick and the proximity signal. We use the same experimental setup as the previous section, but place the stick at different locations relative to the piezo elements. We denote location 0 as a 0 cm horizontal offset from the receiver piezo element (the rod is pointed directly at the receiver). The piezo receiver and transmitter are 20 cm horizontally apart. We performed the approach experiment 13 times at each location. The results in Figure~\ref{fig:d_l} show that \systemname can detect the approaching rod at all approach locations tested. The maximum detection distances are slightly shorter for starting positions closer to the transmitter, with the shortest range being 12.9 cm at the greatest distance from the receiver, compared to a maximum range of 18.3 cm directly over the receiver. 

\subsubsection{Detecting proximity vs. dummy robot material} While the Kinova Jaco manipulator used in \systemname is made of carbon fiber composite, other robot arms might be made of different materials. To study the effect robot material has on the LSW proximity signal, we attach the same piezo pair to 17 different ``robots'' made of different materials, which are shown in Figure~\ref{fig:dummy}. The materials used include brass, aluminum pipe in different shapes, galvanized steel,
Polyvinyl chloride (PVC), and wooden pipe. Again we use the stick apparatus to approach the dummy robot surfaces. The rod approaches vertically toward the midpoint between the transmitter and receiver pair. We perform the experiment 13 times for each dummy robot. Figure~\ref{fig:d_o} shows the maximum detection distances for each dummy robot. The proximity signal can not be detected for material 4 (a floppy thin foil), 7 (a stick), and 12 (a wooden plate). The PVC pipe (material 2) has a notably shorter detection distance (6.5 cm). The other materials we tested produce an LSW signal, with a maximum detection distance among all materials of 37.3 cm from a thin aluminum plate (material 13). In practice, most metals and rigid plastics allow LSW sensing at a reasonable range, as does the carbon fiber composite on the robot. 

\subsubsection{Speed limitations} Given a maximum detection distance $D$ and a system response time $t_d$ (from both mechanical delay and system processing delay), we can calculate an upper approach speed limit $v_{max}$ for \systemname. This speed is the highest relative movement speed between the robot and the obstacle that \systemname can respond to prior to impact, compared as $v_{max}=\frac{D}{t_d}$. The system response time includes everything up to the measured full stop of the robot, $t_d$, is around $150$ ms. As can be seen from the previous micro benchmark studies, \systemname has a maximum detection range on the Kinova Jaco of 18.3 cm. Therefore the upper speed limit for our testbed is 122 cm/s, which is well above the maximum end effector speed of our robot arm (20 cm/s), though fast-moving obstacles could exceed this limit, and adversarial approach vectors could reduce it to 72 cm/s based on our tests. Future optimization of the system response time and maximum detection range could be performed to raise the speed limit as well.

\subsection{Field Study}
\label{subsec:on_robot}

To evaluate the practical effectiveness of our proposed system, we performed several experiments testing \systemname on a Kinova Jaco manipulator to assess the performance of \systemname in a realistic setting. We tested \systemname on 24 objects with various dielectrics, including a human hand, which are shown in Figure~\ref{fig:ob}. In each trial, the object approaches and collides with the arm- the objective of \systemname is to detect the object in the time window 0.5 to 0.2 seconds before impact, which allows the arm enough time to react by stopping in place.
We recorded more than 2000 real world on-robot approach trials with these objects. We hit the stationary arm 1152 times; we programmed the arm to hit a stationary object, and performed that experiment 120 times; we hit the moving arm with an object 1152 times. In each case, the data was collected across two days to include variation in background noise. 
Because our measured system response time $t_d$ is around $150$ ms, we truncate the 0.5-0.2 second window before collision, and consider that window as one datapoint. We also collected negative datapoints in 0.3 second windows where the arm is stationary or randomly moves with no obstacles in proximity.


\subsubsection{Static Robot, Moving Object}
\label{subsubsec:robot_static}
First, we tested the ability of \systemname mounted on a stationary arm to detect approaching objects. This test case avoids many of the challenges described in Section \ref{subsec:lsw_robot}, such as channel disturbance from robot motion, and self-detection. We approached the stationary arm with all 24 objects with diverse angles, approach locations, and approach speeds deliberately introduced by a human operator. We divided the 1152 positive datapoints in half based on the day it was collected, and used one day for training and the other for testing. 
As this experiment lacks complex non-linear noise factors, we used an SVM with an RBF kernel for the classification task. This SVM model obtained both a true positive rate (TPR) and a true negative rate (TNR) of 100\% on the test data. 

\begin{table}[t]
  \centering
    \begin{tabular}{|c|c|c|}
        \hline
         \textbf{Static Object TPR} & \textbf{Moving Object TPR} & \textbf{TNR} \\\hline
        100\% & 95.3\% & 99.1\% \\\hline
    \end{tabular}
\caption{\textbf{Final TPR and TNR for static and mobile objects with the CNN classifier.} The robot arm is moving for both static and mobile object trials.}
\label{tab:tw}
\vspace{-0.2in}
\end{table}

\begin{table*}[t]
    \centering
    \begin{tabular}{|c|c|c|c|c|c|c|c|c|c|c|c|c|}
        \hline
         & Book end & Keyboard box & Transformer & Notebook & PVC pipe & Steel pipe & Plastic box & Sound absorber\\\hline 
        \textbf{TPR} & 93.8\%& 95.8\%& 91.7\%& 91.7\%& 97.9\%& 95.8\%& 93.8\%& 97.9\%\\\hline
        & Alum. box & Acid battery & Stick & Human hand & Cellphone & Cardboard box & Paper towels & Rope bundle \\\hline
        \textbf{TPR} & 97.9\%& 93.8\%& 93.8\%& 97.9\% & 93.8\% & 97.9\% & 97.9\%& 93.8\%\\\hline
         & Painted pipe & Foil Duct & Alum. pipe & Alum. plate & Headphones & Box w/contents & Ventilation duct & Wooden statue \\\hline
        \textbf{TPR}& 91.7\%& 100\%& 95.8\%& 93.8\%& 100\%& 100\%& 97.9\%&91.7\%\\\hline
    \end{tabular}
    \caption{Per-object results from our large scale on-robot experiments using the CNN classifier. }
\vspace{-.2cm}
\label{tab:pob}
\vspace{-0.2in}
\end{table*}

\subsubsection{Moving Robot, Static Object}
\label{subsubsec:hit_static}
For our next set of experiments, we considered the opposite case, in which the robot moves while the obstacle remains static. Accurate detection in this case is more difficult than the previous case, as robot motion introduces complex noise signals as discussed in Section \ref{subsec:lsw_robot}. To test this case, we had the robot move on a repeatable trajectory which intersects with a cardboard box suspended in the air by a string. 
In this case, the 0.3 second window SVM classifier performs much worse, and only achieves an inter-day generalization accuracy of 75.8\%, due to the variation in the signal introduced by robot noise and movement-induced self-detections.

In response, we trained a 1D CNN classifier on this data as detailed in Section~\ref{subsec:classification}. 
This classifier achieves a 0.01 second window classification accuracy of 96.7\%. This test shows that the CNN classifier can handle the self-detections and channel disturbances present in real-world robotic LSW data, even on smaller datasets (only 60 positive trials were used for training, with 60 for testing).

\subsubsection{Moving Robot, Moving Object}
\label{subsubsec:hit_moving}
Finally, we evaluated \systemname on the most complex and realistic case, where both static and moving objects are present.
We collected a much larger dataset of moving robot/moving object trials, consisting of 24 trials of each of the 24 objects for two separate days, or 1152 positive trials in all. In each trial, the robot executes a random movement trajectory while a human operator taps the robot with the corresponding object, held in their hand, making an effort to sample a diversity of approach angles, speeds, and impact locations. 

We trained our CNN classifier on this data
with half the objects used for training and half for testing. We also included the static object data from the preceding section so that our final classifier can detect both static and moving obstacles.
Overall classifier accuracy was 83.9\%, lower than the static-only experiment due to the diversity of robot movements and object approach directions. However, when we apply the sliding window detector described in Section \ref{subsec:classification} to the classifier's predictions, we get moving and static object detection TPR's of 95.3\% and 100\% respectively, with a combined TNR of 99.1\% as shown in Table~\ref{tab:tw}. The receiver operating curve (ROC) for this detector is shown in Figure~\ref{fig:roc}. We selected a positive detection threshold of 0.717 by estimating the inflection point of the combined ROC curve, then used that threshold to compute the individual values. We note that as the drawbacks of false positives are small (the robot halts for a brief period), TPR's could be increased by selecting a lower threshold at the cost of more halts.
We also show the per-object TPR breakdown in Table \ref{tab:pob} (The training and test split was reversed to get test performance for all objects). While there is inter-object variation, \systemname achieves $>$91\% TPR for all objects.  This result shows that \systemname is capable of performing proximity detection in a real-world setting with high accuracy, and that sliding window detection greatly improves system performance by avoiding transient misclassifications.
\vspace{-0.1in}

\section{Conclusion}
\label{s:concl}
\vspace{-0.04in}
In this paper, we presented 
\systemname, a full-surface proximity detection system mounted on a robot arm segment. \systemname employs the LSW generated by a pair of piezo elements to enable no-dead-spot proximity detection for robot collision avoidance. \systemname obtains $>$95\% TPR and $>$99\% TNR in realistic on-robot experiments. \systemname is low cost and lightweight, and can be deployed on other commercial off the shelf robots with minimal modification. 

In the future, we decide to further boost the robustness of the system in order to enable the robot implementing more challenging tasks while performing collision avoidance. We plan to address the robot movement interference mentioned in Section~\ref{subsec:lsw_robot} by exploring more advanced signal processing techniques. We also plan to explore further applications of the LSW, such as tactile sensing and object recognition.
\vspace{-0.05in}

\bibliographystyle{IEEEtran}
\bibliography{references}

\end{document}